\documentclass{article}

\usepackage[preprint]{neurips_data_2023}

\usepackage[utf8]{inputenc} % allow utf-8 input
\usepackage[T1]{fontenc}    % use 8-bit T1 fonts
\usepackage{pifont}         % for checkmarks and crosses
\usepackage{hyperref}       % hyperlinks
\usepackage{url}            % simple URL typesetting
\usepackage{booktabs}       % professional-quality tables
\usepackage{amsfonts}       % blackboard math symbols
\usepackage{nicefrac}       % compact symbols for 1/2, etc.
\usepackage{microtype}      % microtypography
\usepackage[dvipsnames,hyperref,table]{xcolor}
\usepackage{amsmath}
\usepackage{amssymb}
\usepackage{subcaption}
\usepackage[font=small]{caption}
\usepackage{mathtools}
\usepackage[algoruled]{algorithm2e}
\usepackage{multirow}
\usepackage{tabularx}
\usepackage{siunitx}
\usepackage{enumitem}
\usepackage{thmtools}
\usepackage{thm-restate}
\usepackage{xspace}
\usepackage{makecell}
\usepackage{adjustbox}
\usepackage{setspace}
\usepackage{placeins}
\usepackage{tocloft}
\usepackage{color, colortbl}
\usepackage{caption}
\usepackage{subcaption}
\usepackage{wrapfig}
\usepackage[square,sort,comma,numbers]{natbib}

\newcommand\sD{\ensuremath{\mathcal{D}}}

\newcommand\dom{d}
\newcommand\Dom{\sD}
\newcommand\Domtrain{\Dom^\mathsf{train}}
\newcommand\Domtest{\Dom^\mathsf{test}}
\newcommand\Dtrain{D^\mathsf{train}}

\newcommand\Dtest{D^\mathsf{test}}

\newcommand\Ptrain{P^\mathsf{train}}
\newcommand\Ptest{P^\mathsf{test}}
\newcommand\Pdom{P_\dom}

\newcommand\qtraindom{q_\dom^\mathsf{train}}
\newcommand\qtestdom{q_\dom^\mathsf{test}}

\newcommand\DGTrain{$train$\xspace}
\newcommand\DGValid{$valid$\xspace}
\newcommand\DGValidIn{$valid$-$in$\xspace}
\newcommand\DGValidOut{$valid$-$out$\xspace}
\newcommand\DGTest{$test$\xspace}
\newcommand\DGTestIn{$test$-$in$\xspace}
\newcommand\DGTestOut{$test$-$out$\xspace}
\newcommand\DVpowerSynt{$VPower_{S}$\xspace}
\newcommand\DVpowerReal{$VPower_{R}$\xspace}
\newcommand\DWeatherData{$Weather$\xspace}

\title{Wild-Tab: A Benchmark For Out-Of-Distribution Generalization In Tabular Regression}

\makeatletter
\newcommand{\printfnsymbol}[1]{%
  \textsuperscript{\@fnsymbol{#1}}%
}
\makeatother

\author{%
  Sergey Kolesnikov \\
  Tinkoff \\
  \texttt{s.s.kolesnikov@tinkoff.ai} \\
}

\begin{document}

\maketitle

\begin{abstract}

Out-of-Distribution (OOD) generalization, a cornerstone for building robust machine learning models capable of handling data diverging from the training set's distribution, is an ongoing challenge in deep learning. While significant progress has been observed in computer vision and natural language processing, its exploration in tabular data, ubiquitous in many industrial applications, remains nascent.
To bridge this gap, we present Wild-Tab, a large-scale benchmark tailored for OOD generalization in tabular regression tasks. The benchmark incorporates 3 industrial datasets sourced from fields like weather prediction and power consumption estimation, providing a challenging testbed for evaluating OOD performance under real-world conditions.
Our extensive experiments, evaluating 10 distinct OOD generalization methods on Wild-Tab, reveal nuanced insights. 
We observe that many of these methods often struggle to maintain high-performance levels on unseen data, with OOD performance showing a marked drop compared to in-distribution performance. 
At the same time, Empirical Risk Minimization (ERM), despite its simplicity, delivers robust performance across all evaluations, rivaling the results of state-of-the-art methods.
Looking forward, we hope that the release of Wild-Tab will facilitate further research on OOD generalization and aid in the deployment of machine learning models in various real-world contexts where handling distribution shifts is a crucial requirement.

\end{abstract}

\section{Introduction}

Tabular data forms the backbone of numerous practical applications across diverse fields such as finance \cite{heaton2018deep, finance2020_2, finance2020_1}, healthcare \cite{health2018, health2017, DBLP:journals/corr/ShickelTBR17}, and e-commerce \cite{10.1145/2988450.2988454, 10.1145/2959100.2959190}. 
Over the past decade, deep learning models have achieved remarkable success in analyzing this type of data, often surpassing traditional statistical methods  \cite{arik2020tabnet, tabular2022, gorishniy2021revisiting, rubachev2022revisiting}. 
However, akin to other machine learning domains, these models tend to overfit to the training distribution, resulting in subpar generalization to out-of-distribution data \cite{gulrajani2020search, koh2021wilds, malinin2021shifts}.
The intricacies of this problem are particularly pronounced in tabular regression tasks, wherein models are tasked with predicting a continuous output variable. In such contexts, even minor inaccuracies in the predicted values can precipitate substantial real-world consequences, encompassing erroneous medical diagnoses, flawed financial decisions, or inaccurate path prediction in autonomous vehicles \cite{badue2019selfdriving, janai2021computer, malinin2021shifts}.

While there have been notable progress in devising innovative techniques for out-of-distribution generalization \cite{wang2022generalizing, Zhou_2022} - such as the introduction of Quantile Risk Minimization (QRM) for Probable Domain Generalization \cite{eastwoodProbableDomainGeneralization2023}, or the use of the information bottleneck constraint in combination with invariance for improving out-of-distribution generalization \cite{ahuja2021invariance} - most existing studies focus primarily on static image-based classification tasks \cite{gulrajani2020search, koh2021wilds}. 
Research examining distributional shifts over time \cite{yao2022wild} or out-of-distribution generalization in time series \cite{WOODS} exist, but work addressing OOD generalization in tabular data and regression tasks remains scant.

This paper attempts to bridge this gap, striving for a more profound understanding of distribution shifts in tabular data. 
The primary contributions of our work are:
\begin{itemize}
    \item The introduction of Wild-Tab, a benchmark incorporating three extensive tabular datasets specifically designed for testing OOD robustness. To our knowledge, this marks the first OOD generalization methods benchmark tailored for tabular data and one of the most comprehensive OOD benchmarks presently available (see \autoref{table:benchmarks}). 
    \item Comprehensive experimentation with these datasets using Empirical Risk Minimization (ERM) and a variety of OOD generalization strategies. Our findings suggest unique challenges and substantial opportunities for enhancement in out-of-distribution generalization in tabular regression, as shown in \autoref{table:gen_gap}.
    \item Evaluation of advanced tabular models, including MLP-PLR and MLP-T-LR~\cite{gorishniy2022embeddings}. Interestingly, the addition of these sophisticated models did not yield the expected performance boost, highlighting the need for further adaptations of these techniques to the OOD setup.
    \item An assessment of the impact of hyperparameter tuning on the performance of OOD generalization methods, employing the EVP~\cite{showyourwork} approach. 
    Our findings suggest that information bottleneck-based methods exhibit a higher degree of sensitivity to hyperparameter configuration, necessitating more careful tuning.
\end{itemize}

\begin{table}[!t]
\centering
\caption{Overview of OOD generalization benchmarks and datasets. Top-3 largest datasets used for comparison.}
\label{table:benchmarks}
\resizebox{1.0\textwidth}{!}{
\begin{tabular}{l|lll|lll|lll}
    \toprule
    Benchmark & \multicolumn{3}{c|}{Wilds~\cite{koh2020wilds}} & \multicolumn{3}{c|}{Wild-Time~\cite{yao2022wild}} & \multicolumn{3}{c}{Wild-Tab (ours)} \\
    Dataset & FMoW & Amazon & Camelyon17 & arXiv & MIMIC-IV & FMoW-Time & \DVpowerSynt & \DVpowerReal & \DWeatherData \\

    \midrule
    Input &  image & text & image & text & ICD9 codes & image & vector & vector & vector\\
    Task & clf. & clf. & clf. & clf. & clf. & clf. & reg. & reg. & reg. \\
    \# Samples & 523,846 & 539,502 & 455,954 & 2,057,952 & 270,617 & 118,886 & 546,543 & 554,642 & 4,367,323 \\
    \bottomrule
\end{tabular}
}
\end{table}

\begin{table}[h]
    \centering
    \caption{Generalization gap between the In-Distribution (ID) and the Out-Of-Distribution (OOD) test performance of ERM on the Wild-Tab benchmark. See Section~\ref{section:BenchmarkResults} for more details.}
    \label{table:gen_gap}
    % \vspace{-13.2pt}
    \begin{minipage}{\linewidth}
        \centering
        \begin{tabular}{lccc}
            \toprule
            \textbf{Dataset} & \multicolumn{2}{c}{\textbf{MAE}} &  \\
            \cmidrule{2-3}\\[-3.4ex]
             & \multirow{2}{*}{ID} & \multirow{2}{*}{OOD} & \multirow{2}{*}{\textbf{Gap (\%)}} \\[-0.7ex]
             & & & \\[-0.3ex]
            \midrule
            \DVpowerSynt & $875_{\pm13}$ & $933_{\pm16}$ & 6.6 \\
            \DVpowerReal & $996_{\pm36}$ & $1577_{\pm133}$ & 58.3 \\
            \DWeatherData & $1.353_{\pm0.024}$ & $1.741_{\pm0.008}$ & 28.6 \\
            \bottomrule
        \end{tabular}
    \end{minipage}
\end{table}

\paragraph{Why OOD Generalization in Tabular Regression?}
In recent years, the deep learning field has witnessed the emergence of a variety of \textit{foundational} models tailored for image and text data \cite{anil2023palm, chowdhery2022palm, openai2023gpt4, radford2021learning, rombach2022highresolution}. 
This resulted in CLIP making significant strides in OOD generalization for static computer vision tasks \cite{cha2022domain}, and ChatGPT demonstrating consistent outperformance in a majority of OOD classification tasks \cite{wang2023robustness}. 
However, the lack of \textit{pretrained} models on tabular data, whether or not they help address OOD generalization, presents a potential hurdle for OOD methods and underscores the need for further research in this direction. 
Our benchmark, Wild-Tab, is envisioned as a contribution to help shed light on this crucial issue.

Another aspect warranting attention in the tabular space is the lack of advanced data \textit{augmentation} techniques \cite{rubachev2022revisiting}. 
Unlike the computer vision field, where augmentations have shown effectiveness in OOD generalization benchmarks \cite{zhong2022adversarial}, the absence of such techniques in the tabular space poses potential challenges. 
These challenges, yet to be fully explored, could drive the development of OOD methods that are inherently resilient to perturbed data.

\clearpage
\section{Problem and Evaluation Settings}
\subsection{Problem Formulation} 
Adopting the established OOD generalization paradigm \cite{koh2020wilds}, we conceive the entire data distribution as an ensemble of $D$ domains, denoted by $\Dom = {1,\dots,D}$. Within this framework, every domain $\dom \in \Dom$ is paired with a data distribution $P_d(X, Y)$, linking input attributes $X$ to their respective targets $Y$. To formulate the distribution shift setting, we introduce the training dataset $\Dtrain$ that adheres to the distribution $\Ptrain = \sum_{\dom\in\Dom} \qtraindom\Pdom$, incorporating blend weights $\qtraindom$ for each domain $\dom$. Conversely, the test dataset, $\Dtest$, aligns with the $\Ptest = \sum_{\dom\in\Dom} \qtestdom\Pdom$ distribution. The primary objective of OOD generalization is to leverage train dataset $\Dtrain$ to derive a model $f$ that is proficient on the test data $\Dtest$. The performance of model $f$ in domain $d$ is quantified via the risk metric $R^{d}(f) = \mathbb E^d\big[\ell (f(X), Y)\big]$, where $\mathbb{E}^d$ represents the expected value over the distribution $P_d$ and $\ell \rightarrow \mathbb{R}_{\geq 0}$ denotes the loss function.

\textbf{Domain Generalization (\textit{source shift})} Within domain generalization, the goal is to expand towards test domains, symbolized as $\Domtest$, that are distinct from the training domains, $\Domtrain$, thus ensuring $\Domtrain \cap \Domtest=\emptyset$.

\textbf{Subpopulation Shift (\textit{temporal shift})} Addressing subpopulation shifts requires a model to maintain consistent performance across various domains encountered during its training. Explicitly, all test domains are recognized during training, illustrated by $\Domtest\subseteq\Domtrain$. Nevertheless, proportions between the domains might experience fluctuations.

\textbf{Hybrid Shift} While the classes of source and temporal shifts offer a structured approach towards OOD shifts, segregating a problem strictly as either remains challenging. Take a weather forecasting task for example. We could have sensors installed on various residences across multiple climate zones aiming to generalize for these zones -- this is a \textit{source} shift. Alternatively, we could have a single sensor on a single residence aiming to generalize for temperature variations throughout the year -- a \textit{temporal} shift. Lastly, we could have sensors across different climate zones aiming to generalize for annual temperature changes -- a \textit{hybrid} scenario.

In this paper, we primarily focus on \textit{hybrid} case, which involves both \textit{source} and \textit{temporal} shifts.
This focus on the hybrid scenario is driven by its common occurrence in real-world datasets, which frequently exhibit both \textit{source} and \textit{temporal} shifts, as discussed in Section~\ref{sec:datasets}

\subsection{Evaluation Strategies}

Evaluating OOD generalization in \textit{hybrid} scenarios involving both \textit{source} and \textit{temporal} shifts presents a unique set of challenges. First, the increased complexity of these scenarios necessitates models that can simultaneously capture variations across different sources and time-points, making the task of ensuring robust generalization more demanding. Second, the uncertainties inherent in both \textit{source} and \textit{temporal} shifts compound in hybrid cases, thereby exacerbating the difficulty of predicting model behavior and performance. Finally, potential interaction effects between source and time domains add an additional layer of complexity. These effects can result in nuanced and complex data distributions that are challenging to model, and their presence can further complicate the evaluation of OOD generalization.

\begin{figure*}[!t]
  \centering
  \makebox[\textwidth][c]{\includegraphics[width=0.8\textwidth]{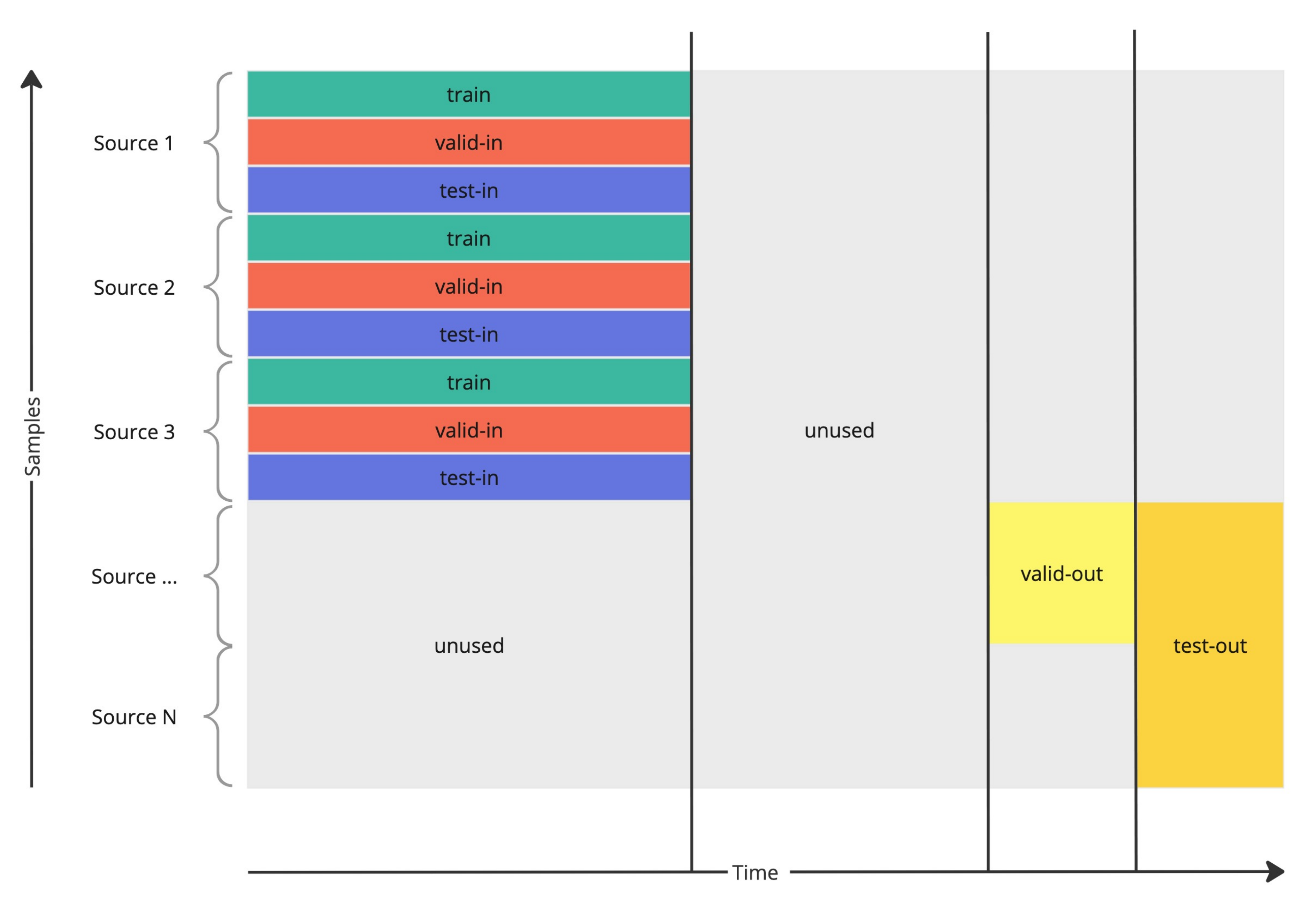}}
  \caption{Hybrid dataset partitioning for out-of-distribution generalization evaluation.}
  \label{fig:train_test}
\end{figure*}

To address the challenges above, the Shifts evaluation protocol~\cite{malinin2021shifts} was introduced. Firstly, a hybrid dataset is temporally partitioned into \DGTrain, \DGValid, and \DGTest parts. After that, it further split into \DGValidIn, \DGValidOut, \DGTestIn, and \DGTestOut based on source domains to represent in-domain and shifted segments, respectively. A visual representation of this partitioning can be seen in \autoref{fig:train_test}. A model is then trained on the \DGTrain set, with the \DGValid set used for model selection and the \DGTest set used for the final performance evaluation. Furthermore, the disparity between $in$ and $out$ performance on \DGValid and \DGTest sets is used as a measure of model generalization.
Throughout our experiments, we employ datasets originating from Shifts~\cite{malinin2022shifts, malinin2021shifts}, thereby upholding the proposed evaluation protocol.

\clearpage
\section{Wild-Tab Benchmark}
\label{section:benchmark}

\subsection{Datasets}
\label{sec:datasets}

\paragraph{Vessel Power Estimation}
We use the Shifts vessel power estimation dataset~\cite{malinin2022shifts} as our first industrial dataset with real-life OOD challenges. 
It requires us to anticipate the energy utilization of cargo ships under various weather and operational conditions. Owing to limited, noisy sensor data, making precise predictions becomes challenging, necessitating resilient power consumption models to curtail costs and CO$_2$ emissions. As per \cite{malinin2022shifts}, we adopt synthetic (\DVpowerSynt) and real (\DVpowerReal) datasets for our analysis, each containing minute-by-minute measurements from shipboard sensors across four years. This results in approximately half a million observations, each with 10 input features and one regression target. 
We use publicly available datasets and canonically repartitioned them along two dimensions: wind speed and time, following \cite{malinin2022shifts} methodology~\footnote{Public access to canonical \DGTest portions of \DVpowerSynt and \DVpowerReal is restricted, necessitating dataset reconfiguration.}.
Further details about these datasets can be found in \autoref{appendix:datasets}.

\paragraph{Tabular Weather Prediction}
The last used dataset, namely the Shifts Weather Prediction dataset~\cite{malinin2021shifts} (\DWeatherData), provides a scalar regression task necessitating the forecast of air temperatures two meters above ground level at specific geographical coordinates and time points. The full dataset comprises more than 4 million instances, including 123 weather features, 4 metadata attributes, and a single regression target, respectively. 
It spans an entire year—from September 1st, 2018, to September 1st, 2019—and encapsulates samples from all climate categories: tropical, dry, mild temperature, snow, and polar. 
We use a canonically partitioned dataset along two dimensions: climate type and time.
More specifics about the \DWeatherData dataset can be found in \autoref{appendix:datasets}.

\subsection{Deep Learning Models for Tabular Data}

We employ MLP as the baseline deep learning model for comparative analysis and study ablation. Adhering to the setup recommendations proposed by \cite{gorishniy2021revisiting}, we utilize regularization techniques such as dropout and weight decay in our models. 
While the primary focus remains the evaluation of existing OOD generalization techniques with basic tabular data setups, we also delve into more complex deep learning models. 
For further analysis, we test MLP models that incorporate numerical feature embeddings~\cite{gorishniy2022embeddings}, namely MLP-PLR and MLP-T-LR and FT-Transformer~\cite{gorishniy2021revisiting}. These models represent the current state-of-the-art in the field of deep learning for tabular data \cite{gorishniy2022embeddings}.
For model optimization, we use the AdamW optimizer \cite{adamw}, while deliberately avoiding learning rate schedules and setting batch sizes as per the respective dataset sizes. More on model specifics can be found in \autoref{appendix:imp_details}.

\subsection{Out-of-Distribution Generalization Techniques}

Numerous techniques have been suggested to train models that exhibit increased resilience to specific distribution shifts, compared to conventional models trained via Empirical Risk Minimization (ERM). These methodologies generally leverage domain annotations during training, with the intention of learning a model that can generalize across domains. 
In our Wild-Tab benchmark, we test 10 OOD generalization methods, suitable for regression tasks, including the ERM baseline:

\begin{itemize}
\item \textbf{CORAL} \cite{sun2016deep} penalizes differences in the means and covariances of the feature distributions.
\item \textbf{DANN} \cite{ganin2016domainadversarial} encourages domain-invariant representations by training a feature extractor and domain classifier adversarially.
\item \textbf{EQRM} \cite{eastwoodProbableDomainGeneralization2023} uses Quantile Risk Minimization to formulate a predictor that minimizes its risk distribution's $\alpha$-quantile across domains.
\item \textbf{Group DRO} \cite{hu2018does,sagawa2020group} uses distributionally robust optimization to explicitly minimize loss on the worst-case domain during training.
\item \textbf{IRM} \cite{arjovsky2019invariant} penalizes feature distributions having different optimal linear heads for each domain.
\item \textbf{IB-IRM} \cite{ahuja2021invariance} introduces an information bottleneck optimization into the \textbf{IRM} method.
\item \textbf{IB-ERM} \cite{ahuja2021invariance} is a simplified version of the \textbf{IB-IRM} model, without the invariance constraint.
\item \textbf{MMD} \cite{chevalley2022invariant} matches the Maximum Mean Discrepancy of feature distributions for each domain.
\item \textbf{VREx} \cite{krueger2021out} applies a penalty to the variance of training risks.
\end{itemize}

\subsection{Experiment Design}

\paragraph{Evaluation Metrics}
The performance and robustness of the models to distribution shifts are evaluated using standard regression metrics: Mean Absolute Error (MAE), Mean Absolute Percentage Error (MAPE), and Root Mean Square Error (RMSE). Lower scores on these metrics on the test sets indicate stronger robustness of the models to the distributional shift. 
Due to paper length restrictions, we focus our comparison on the MAE metric, with additional metrics' results available in \autoref{appendix:extra_exps}.

\paragraph{Hyperparameter Optimization}
We adopt the DomainBed workflow \cite{gulrajani2020search} for hyperparameter search and model selection, which enables a comprehensive and unbiased assessment of OOD generalization techniques. We conduct a random search over 20 hyperparameter configurations, replicated thrice for error estimation. 
The performance of the model chosen with the model selection method is subsequently reported.
More on the framework and search spaces for hyperparameters can be found in \autoref{appendix:imp_details}.

\paragraph{Model Selection}
In alignment with \cite{WOODS, liang2022metashift, malinin2021shifts}, we employ the following model selection procedures. Given the canonically partitioned datasets, we select the model based on four possible performance measurements on validation: average-in-domain, average-out-domain, worst-in-domain, and worst-out-domain. Due to paper length constraints, our method comparison uses average-out-domain validation model selection, as suggested by WILDS~\cite{koh2020wilds}. Additional results are available in the \autoref{appendix:extra_exps}.

\newpage
\section{Experiments}
\label{section:Experiments}

% \subsection{OOD generalization algorithms results}
\subsection{Benchmark Results}
\label{section:BenchmarkResults}

The experimental results of the Wild-Tab benchmark are outlined in \autoref{table:BenchmarkResults}.
Each dataset's performance is measured via the MAE, expressed in kW for \DVpowerSynt and \DVpowerReal datasets and in Celsius for the \DWeatherData dataset. Supplementary results using RMSE and MAPE metrics are available in \autoref{appendix:extra_exps}. In all scenarios, smaller figures denote superior outcomes. The findings are sorted based on the data type (validation or test) and partition (in-distribution, out-of-distribution). Several key insights emerge from this examination.

\begin{table}[htbp!]
\centering
\caption{OOD generalization methods performance on Wild-Tab benchmark. 
% Mean MAE $\pm~\sigma$ is quoted for in- and out-domain parts.
}
\label{table:BenchmarkResults}
\resizebox{0.99\textwidth}{!}{
\begin{tabular}{l|l|ll|ll|ll}
\toprule
\multirow{2}*{Data} & \multirow{2}*{Objective} & \multicolumn{2}{c|}{\DVpowerSynt} & \multicolumn{2}{c|}{\DVpowerReal} & \multicolumn{2}{c}{\DWeatherData} \\
& & In & Out & In & Out & In & Out \\
\midrule

\multirow{10}*{Valid} & CORAL & $862_{\pm9}$ & $700_{\pm5}$ & $1008_{\pm35}$ & $618_{\pm33}$ & $1.35_{\pm0.018}$ & $1.604_{\pm0.013}$ \\
& DANN & $988_{\pm77}$ & $704_{\pm7}$ & $973_{\pm5}$ & $632_{\pm21}$ & $1.474_{\pm0.011}$ & $1.604_{\pm0.009}$ \\
& EQRM & $878_{\pm14}$ & $705_{\pm9}$ & $937_{\pm33}$ & $620_{\pm29}$ & $1.457_{\pm0.038}$ & $1.61_{\pm0.005}$ \\
& ERM & $872_{\pm11}$ & $705_{\pm6}$ & $1002_{\pm36}$ & $628_{\pm10}$ & $1.357_{\pm0.026}$ & $1.603_{\pm0.008}$ \\
& GroupDRO & $873_{\pm18}$ & $705_{\pm3}$ & $1000_{\pm47}$ & $635_{\pm44}$ & $1.36_{\pm0.03}$ & $1.599_{\pm0.004}$ \\
& IB\_ERM & $900_{\pm22}$ & $716_{\pm7}$ & $932_{\pm15}$ & $610_{\pm5}$ & $1.397_{\pm0.008}$ & $1.62_{\pm0.007}$ \\
& IB\_IRM & $860_{\pm3}$ & $714_{\pm4}$ & $919_{\pm16}$ & $632_{\pm53}$ & $1.418_{\pm0.025}$ & $1.623_{\pm0.008}$ \\
& IRM & $869_{\pm8}$ & $704_{\pm2}$ & $1032_{\pm16}$ & $673_{\pm56}$ & $1.352_{\pm0.048}$ & $1.603_{\pm0.002}$ \\
& MMD & $865_{\pm14}$ & $700_{\pm4}$ & $979_{\pm22}$ & $639_{\pm18}$ & $1.371_{\pm0.008}$ & $1.607_{\pm0.008}$ \\
& VREx & $937_{\pm21}$ & $709_{\pm8}$ & $950_{\pm18}$ & $647_{\pm8}$ & $1.355_{\pm0.023}$ & $1.601_{\pm0.001}$ \\
\midrule
\multirow{10}*{Test} & CORAL & $865_{\pm10}$ & $916_{\pm8}$ & $1001_{\pm33}$ & $1679_{\pm184}$ & $1.345_{\pm0.019}$ & $1.741_{\pm0.005}$ \\
& DANN & $993_{\pm77}$ & $1029_{\pm63}$ & $965_{\pm2}$ & $1714_{\pm78}$ & $1.471_{\pm0.01}$ & $1.77_{\pm0.014}$ \\
& EQRM & $881_{\pm14}$ & $950_{\pm12}$ & $931_{\pm34}$ & $1627_{\pm149}$ & $1.455_{\pm0.036}$ & $1.761_{\pm0.003}$ \\
& ERM & $875_{\pm12}$ & $932_{\pm16}$ & $996_{\pm35}$ & $1576_{\pm133}$ & $1.353_{\pm0.024}$ & $1.741_{\pm0.008}$ \\
& GroupDRO & $875_{\pm17}$ & $934_{\pm15}$ & $993_{\pm46}$ & $1784_{\pm63}$ & $1.356_{\pm0.029}$ & $1.734_{\pm0.016}$ \\
& IB\_ERM & $904_{\pm20}$ & $1105_{\pm60}$ & $927_{\pm15}$ & $1653_{\pm33}$ & $1.398_{\pm0.008}$ & $1.742_{\pm0.007}$ \\
& IB\_IRM & $862_{\pm5}$ & $1136_{\pm99}$ & $911_{\pm17}$ & $1531_{\pm32}$ & $1.418_{\pm0.026}$ & $1.759_{\pm0.003}$ \\
& IRM & $871_{\pm10}$ & $934_{\pm20}$ & $1024_{\pm14}$ & $1782_{\pm168}$ & $1.348_{\pm0.048}$ & $1.737_{\pm0.016}$ \\
& MMD & $868_{\pm14}$ & $917_{\pm0}$ & $973_{\pm23}$ & $1579_{\pm58}$ & $1.367_{\pm0.007}$ & $1.737_{\pm0.01}$ \\
& VREx & $942_{\pm22}$ & $1094_{\pm99}$ & $946_{\pm17}$ & $1628_{\pm49}$ & $1.349_{\pm0.02}$ & $1.739_{\pm0.003}$ \\

\midrule
\end{tabular}
}
\end{table}

\paragraph{Benchmarked datasets have a significant generalization gap}
It can be observed that the performance of the algorithms varies considerably between the in-distribution and out-of-distribution settings. Notably, the MAE
% , RMSE, and MAPE 
values are consistently higher for the out-of-distribution test set compared to the in-distribution test set. This suggests that the algorithms struggle to generalize well to unseen data, highlighting the presence of a substantial generalization gap in the benchmarked datasets.

\paragraph{Validation performance not sufficient}
Although validation performance often provides a reliable forecast of in-distribution test outcomes,
% (IB\_IRM on \DVpowerReal or CORAL on \DWeatherData)
we find this is not always true for OOD test results. 
Multiple algorithms 
% (EB\_IRM on \DVpowerReal or GroupDRO on \DWeatherData) 
that exhibit strong validation performance on OOD data do not necessarily duplicate this success on OOD test data, reaffirming the conclusion that sound validation performance does not ensure robust OOD generalization.

\paragraph{When all conditions are equal, no algorithm outperforms ERM by a significant margin}
We observe that ERM's performance is on par with other OOD generalization techniques on the Wild-Tab benchmark. The average MAE for ERM on out-of-distribution data, although not necessarily the smallest, aligns with the overall range of other methods, signifying that ERM remains a solid choice for OOD generalization in tabular regression tasks. We do not assert that these algorithms cannot surpass ERM, but rather we stress the necessity for innovative strategies or substantial revisions to existing methods to significantly outperform ERM.

\clearpage
\subsection{Analysis}

\paragraph{Impact of Model Architecture}
In examining the influence of model architecture on OOD generalization, we benchmark MLP models with numerical feature embeddings, specifically MLP-PLR, MLP-T-LP~\cite{gorishniy2022embeddings} and FT-Transformer~\cite{gorishniy2021revisiting}, as detailed in \autoref{table:ArchResults}. While we anticipated a significant improvement with the introduction of more complex architectures, the actual outcomes suggest a contrary trend. Notably, advanced configurations MLP-PLR and MLP-T-LR across various objectives appear to impair the out-of-distribution generalization performance compared to their simpler MLP counterpart.
Similarly, for the FT-Transformer, while there's a discernible boost in validation OOD performance, the corresponding test performance exhibits heightened instability.

\begin{table}[htbp!]
\centering
\caption{OOD generalization methods performance on \DVpowerReal dataset with varied model architectures.}
\label{table:ArchResults}
\resizebox{1.0\textwidth}{!}{
\begin{tabular}{l|l|ll|ll|ll|ll}
\toprule
\multirow{2}*{Data} & \multirow{2}*{Objective} & \multicolumn{2}{c|}{MLP} & \multicolumn{2}{c|}{MLP-PLR} & \multicolumn{2}{c}{MLP-T-LR} & \multicolumn{2}{c}{FT-Transformer} \\
& & In & Out & In & Out & In & Out & In & Out \\
\midrule

\multirow{10}*{Valid} & CORAL & $1008_{\pm35}$ & $618_{\pm33}$ & $789_{\pm32}$ & $572_{\pm24}$ & $698_{\pm60}$ & $520_{\pm25}$ & $1097_{\pm150}$ & $586_{\pm9}$ \\
& DANN & $973_{\pm5}$ & $632_{\pm21}$ & $889_{\pm15}$ & $594_{\pm60}$ & $1432_{\pm98}$ & $565_{\pm50}$ & $1010_{\pm49}$ & $629_{\pm18}$ \\
& EQRM & $937_{\pm33}$ & $620_{\pm29}$ & $891_{\pm168}$ & $568_{\pm30}$ & $722_{\pm47}$ & $572_{\pm87}$ & $1064_{\pm37}$ & $592_{\pm28}$ \\
& ERM & $1002_{\pm36}$ & $628_{\pm10}$ & $887_{\pm110}$ & $571_{\pm19}$ & $1282_{\pm81}$ & $568_{\pm74}$ & $1048_{\pm24}$ & $582_{\pm8}$ \\
& GroupDRO & $1000_{\pm47}$ & $635_{\pm44}$ & $769_{\pm56}$ & $578_{\pm13}$ & $845_{\pm127}$ & $576_{\pm16}$ & $1025_{\pm74}$ & $604_{\pm5}$ \\
& IB\_ERM & $932_{\pm15}$ & $610_{\pm5}$ & $918_{\pm146}$ & $618_{\pm4}$ & $747_{\pm53}$ & $573_{\pm19}$ & $984_{\pm21}$ & $589_{\pm21}$ \\
& IB\_IRM & $919_{\pm16}$ & $632_{\pm53}$ & $926_{\pm124}$ & $618_{\pm19}$ & $827_{\pm23}$ & $630_{\pm15}$ & $1041_{\pm60}$ & $621_{\pm6}$ \\
& IRM & $1032_{\pm16}$ & $673_{\pm56}$ & $853_{\pm38}$ & $620_{\pm34}$ & $833_{\pm60}$ & $597_{\pm16}$ & $1008_{\pm38}$ & $575_{\pm9}$ \\
& MMD & $979_{\pm22}$ & $639_{\pm18}$ & $843_{\pm30}$ & $587_{\pm54}$ & $782_{\pm56}$ & $600_{\pm51}$ & $1008_{\pm69}$ & $588_{\pm18}$ \\
& VREx & $950_{\pm18}$ & $647_{\pm8}$ & $778_{\pm67}$ & $564_{\pm49}$ & $1596_{\pm64}$ & $583_{\pm40}$ & $983_{\pm32}$ & $599_{\pm31}$ \\
\midrule
\multirow{10}*{Test} & CORAL & $1001_{\pm33}$ & $1679_{\pm184}$ & $783_{\pm33}$ & $2327_{\pm154}$ & $690_{\pm61}$ & $1887_{\pm250}$ & $1090_{\pm147}$ & $1948_{\pm377}$ \\
& DANN & $965_{\pm2}$ & $1714_{\pm78}$ & $879_{\pm16}$ & $2149_{\pm266}$ & $1420_{\pm98}$ & $2031_{\pm364}$ & $1005_{\pm52}$ & $2009_{\pm96}$ \\
& EQRM & $931_{\pm34}$ & $1627_{\pm149}$ & $882_{\pm166}$ & $1941_{\pm109}$ & $717_{\pm44}$ & $1898_{\pm376}$ & $1058_{\pm39}$ & $1878_{\pm44}$ \\
& ERM & $996_{\pm35}$ & $1576_{\pm133}$ & $879_{\pm110}$ & $2215_{\pm158}$ & $1272_{\pm84}$ & $2201_{\pm269}$ & $1046_{\pm24}$ & $1883_{\pm130}$ \\
& GroupDRO & $993_{\pm46}$ & $1784_{\pm63}$ & $765_{\pm58}$ & $2437_{\pm302}$ & $837_{\pm127}$ & $2611_{\pm305}$ & $1020_{\pm73}$ & $2009_{\pm93}$ \\
& IB\_ERM & $927_{\pm15}$ & $1653_{\pm33}$ & $913_{\pm142}$ & $1910_{\pm380}$ & $741_{\pm52}$ & $2208_{\pm13}$ & $982_{\pm23}$ & $1877_{\pm121}$ \\
& IB\_IRM & $911_{\pm17}$ & $1531_{\pm32}$ & $919_{\pm125}$ & $2183_{\pm402}$ & $820_{\pm23}$ & $1685_{\pm32}$ & $1039_{\pm58}$ & $1534_{\pm190}$ \\
& IRM & $1024_{\pm14}$ & $1782_{\pm168}$ & $841_{\pm37}$ & $2108_{\pm195}$ & $833_{\pm59}$ & $1948_{\pm55}$ & $1001_{\pm41}$ & $1789_{\pm264}$ \\
& MMD & $973_{\pm23}$ & $1579_{\pm58}$ & $837_{\pm34}$ & $1945_{\pm166}$ & $782_{\pm57}$ & $2041_{\pm245}$ & $1003_{\pm69}$ & $1722_{\pm188}$ \\
& VREx & $946_{\pm17}$ & $1628_{\pm49}$ & $771_{\pm67}$ & $2054_{\pm436}$ & $1587_{\pm65}$ & $2856_{\pm109}$ & $979_{\pm32}$ & $1653_{\pm101}$ \\

\midrule

\end{tabular}
}
\end{table}

\paragraph{Effect of Model Selection Procedure}
We also explore the potential effects of the model selection process on OOD performance. Our benchmark comparisons employ the WILDS methodology, utilizing average-out-domain validation, but we also offer an additional Wild-Tab benchmark following the average-in-domain validation paradigm in \autoref{table:BenchmarkResults2}, thereby limiting model access to OOD data during training. 
Even though minor variances in results and model rankings were observed, the data reinforced consistent trends noted previously: (1) a discernible generalization gap, (2) inconsistent test performance, and (3) marginal improvement over the ERM baseline. A closer look at the \DWeatherData dataset reveals that the ERM exhibits marginally superior outcomes under average-out-domain validation, thus underscoring the value of out-of-distribution validation for OOD generalization.

\begin{table}[htbp!]
\centering
\caption{OOD generalization methods performance on Wild-Tab benchmark with average-in-domain validation.}
\label{table:BenchmarkResults2}
\resizebox{1.0\textwidth}{!}{
\begin{tabular}{l|l|ll|ll|ll}
\toprule
\multirow{2}*{Data} & \multirow{2}*{Objective} & \multicolumn{2}{c|}{\DVpowerSynt} & \multicolumn{2}{c|}{\DVpowerReal} & \multicolumn{2}{c}{\DWeatherData} \\
& & In & Out & In & Out & In & Out \\
\midrule

\multirow{10}*{Valid} & CORAL & $795_{\pm2}$ & $804_{\pm30}$ & $624_{\pm10}$ & $955_{\pm106}$ & $1.293_{\pm0.003}$ & $1.684_{\pm0.017}$ \\
& DANN & $819_{\pm3}$ & $751_{\pm25}$ & $766_{\pm16}$ & $925_{\pm115}$ & $1.325_{\pm0.004}$ & $1.636_{\pm0.016}$ \\
& EQRM & $805_{\pm3}$ & $788_{\pm17}$ & $647_{\pm10}$ & $939_{\pm219}$ & $1.305_{\pm0.003}$ & $1.675_{\pm0.022}$ \\
& ERM & $797_{\pm1}$ & $751_{\pm22}$ & $624_{\pm27}$ & $901_{\pm49}$ & $1.295_{\pm0.005}$ & $1.674_{\pm0.017}$ \\
& GroupDRO & $798_{\pm0}$ & $760_{\pm18}$ & $611_{\pm6}$ & $1235_{\pm170}$ & $1.29_{\pm0.002}$ & $1.704_{\pm0.008}$ \\
& IB\_ERM & $805_{\pm4}$ & $832_{\pm87}$ & $747_{\pm21}$ & $917_{\pm81}$ & $1.34_{\pm0.001}$ & $1.804_{\pm0.004}$ \\
& IB\_IRM & $827_{\pm7}$ & $791_{\pm34}$ & $757_{\pm25}$ & $796_{\pm19}$ & $1.405_{\pm0.003}$ & $1.636_{\pm0.018}$ \\
& IRM & $793_{\pm7}$ & $760_{\pm25}$ & $702_{\pm15}$ & $980_{\pm148}$ & $1.291_{\pm0.003}$ & $1.662_{\pm0.018}$ \\
& MMD & $793_{\pm7}$ & $740_{\pm24}$ & $681_{\pm17}$ & $2109_{\pm841}$ & $1.295_{\pm0.002}$ & $1.671_{\pm0.009}$ \\
& VREx & $800_{\pm1}$ & $754_{\pm27}$ & $697_{\pm9}$ & $1057_{\pm355}$ & $1.293_{\pm0.007}$ & $1.654_{\pm0.013}$ \\
\midrule
\multirow{10}*{Test} & CORAL & $796_{\pm2}$ & $1050_{\pm44}$ & $619_{\pm9}$ & $2354_{\pm123}$ & $1.288_{\pm0.003}$ & $1.811_{\pm0.004}$ \\
& DANN & $822_{\pm5}$ & $1042_{\pm16}$ & $760_{\pm17}$ & $1937_{\pm58}$ & $1.321_{\pm0.002}$ & $1.764_{\pm0.02}$ \\
& EQRM & $806_{\pm2}$ & $1098_{\pm59}$ & $640_{\pm11}$ & $2188_{\pm239}$ & $1.304_{\pm0.001}$ & $1.791_{\pm0.004}$ \\
& ERM & $798_{\pm2}$ & $1022_{\pm12}$ & $619_{\pm27}$ & $2253_{\pm376}$ & $1.291_{\pm0.003}$ & $1.803_{\pm0.007}$ \\
& GroupDRO & $799_{\pm1}$ & $1000_{\pm46}$ & $605_{\pm7}$ & $2400_{\pm297}$ & $1.292_{\pm0.004}$ & $1.837_{\pm0.021}$ \\
& IB\_ERM & $807_{\pm4}$ & $1085_{\pm88}$ & $741_{\pm20}$ & $2146_{\pm151}$ & $1.339_{\pm0.001}$ & $1.925_{\pm0.008}$ \\
& IB\_IRM & $828_{\pm6}$ & $1097_{\pm38}$ & $752_{\pm24}$ & $2083_{\pm148}$ & $1.403_{\pm0.003}$ & $1.768_{\pm0.015}$ \\
& IRM & $795_{\pm7}$ & $1039_{\pm29}$ & $697_{\pm18}$ & $2282_{\pm304}$ & $1.288_{\pm0.0}$ & $1.793_{\pm0.017}$ \\
& MMD & $794_{\pm8}$ & $1047_{\pm90}$ & $677_{\pm16}$ & $2472_{\pm252}$ & $1.292_{\pm0.004}$ & $1.809_{\pm0.022}$ \\
& VREx & $802_{\pm2}$ & $968_{\pm14}$ & $691_{\pm12}$ & $2300_{\pm355}$ & $1.29_{\pm0.007}$ & $1.789_{\pm0.02}$ \\

\midrule

\end{tabular}
}

\end{table}

\begin{figure}
     \centering
     \begin{subfigure}[b]{0.3\textwidth}
         \centering
         \includegraphics[width=\textwidth]{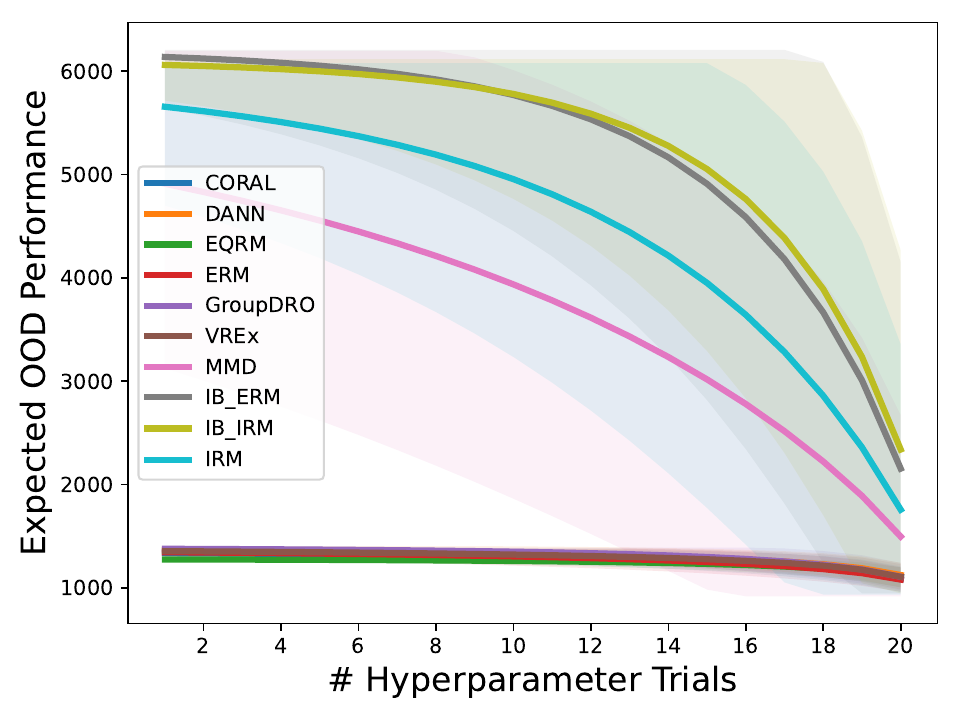}
         % \caption{\DVpowerSynt}
         % \label{fig:y equals x}
     \end{subfigure}
     \hfill
     \begin{subfigure}[b]{0.3\textwidth}
         \centering
         \includegraphics[width=\textwidth]{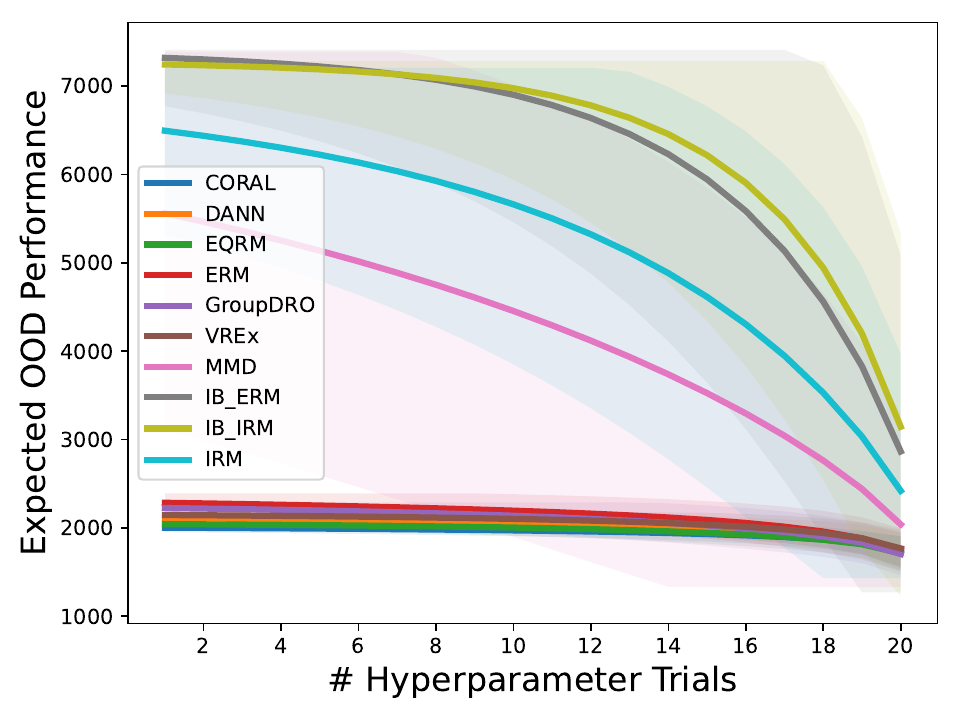}
         % \caption{\DVpowerReal}
         % \label{fig:three sin x}
     \end{subfigure}
     \hfill
     \begin{subfigure}[b]{0.3\textwidth}
         \centering
         \includegraphics[width=\textwidth]{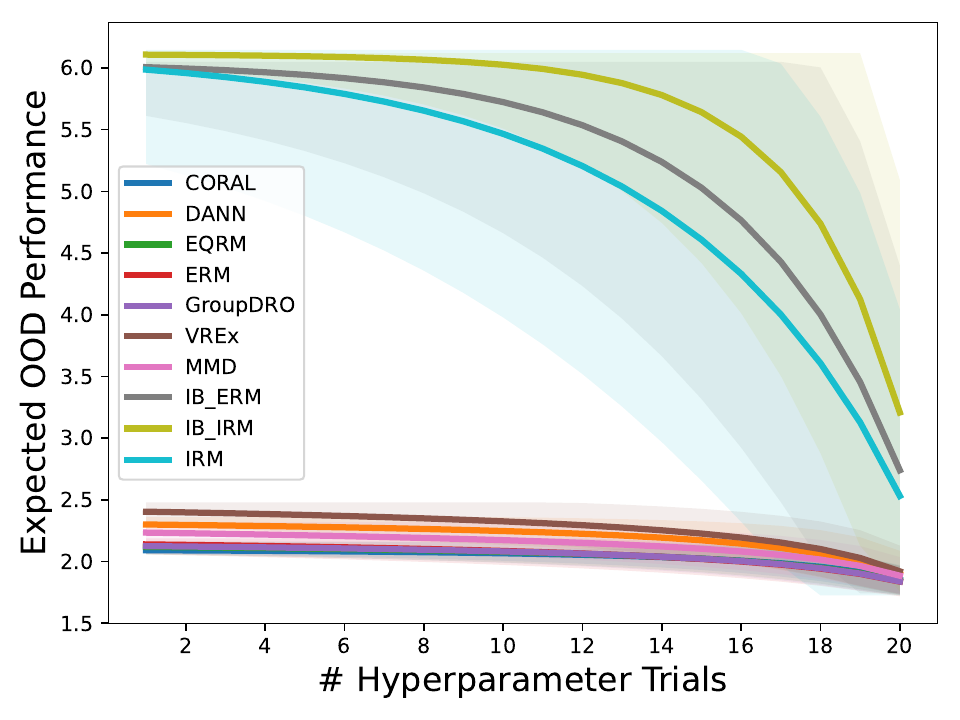}
         % \caption{\DWeatherData}
         % \label{fig:five over x}
     \end{subfigure}
     \begin{subfigure}[b]{0.3\textwidth}
         \centering
         \includegraphics[width=\textwidth]{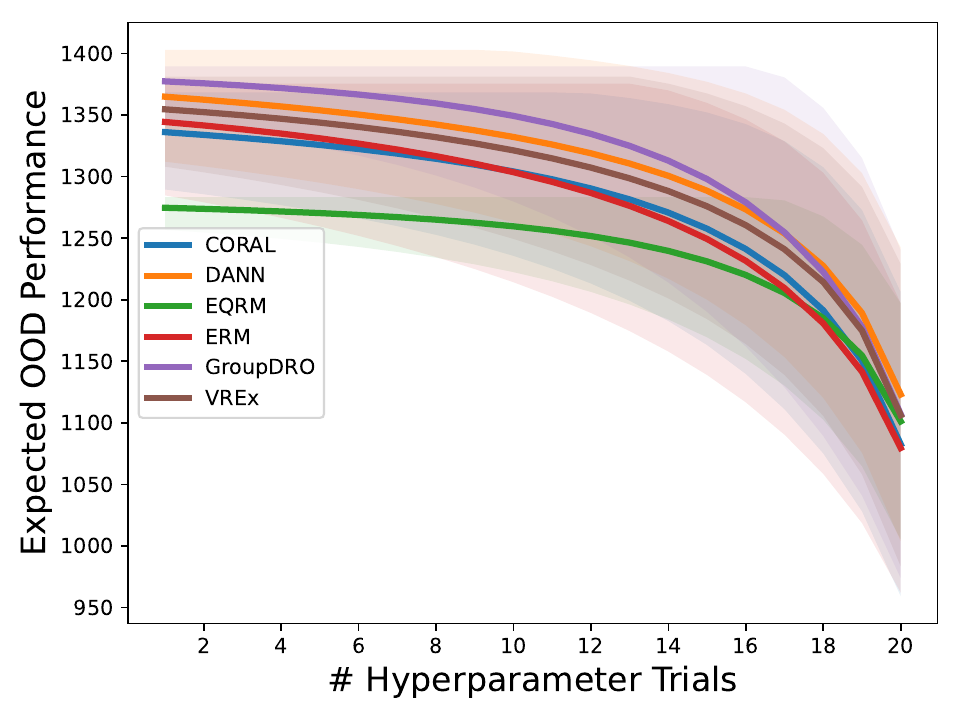}
         \caption{\DVpowerSynt}
         % \label{fig:y equals x}
     \end{subfigure}
     \hfill
     \begin{subfigure}[b]{0.3\textwidth}
         \centering
         \includegraphics[width=\textwidth]{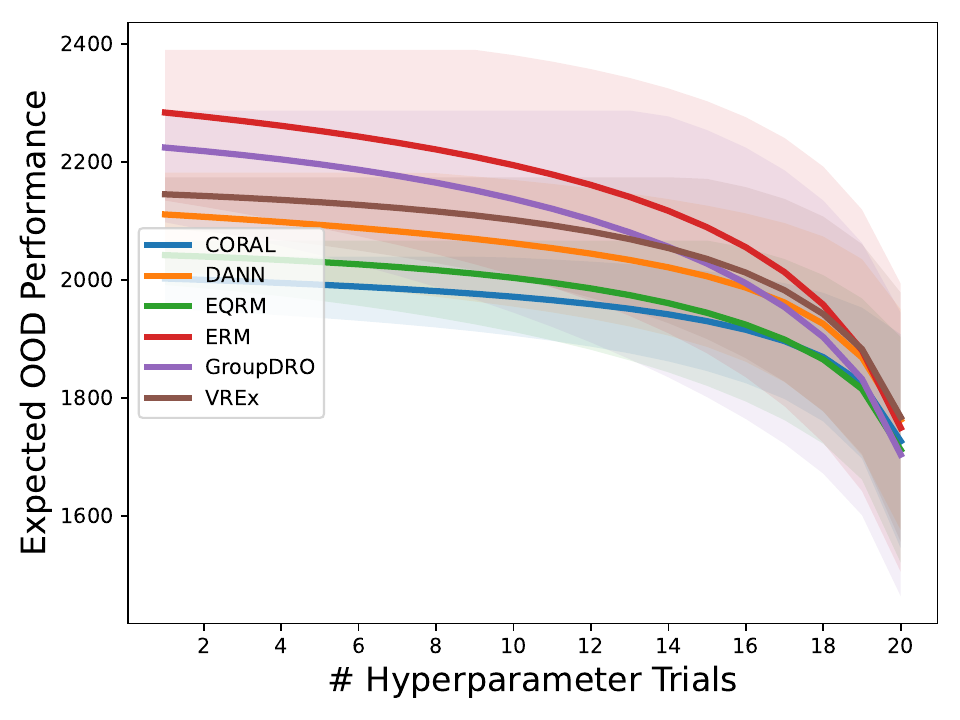}
         \caption{\DVpowerReal}
         % \label{fig:three sin x}
     \end{subfigure}
     \hfill
     \begin{subfigure}[b]{0.3\textwidth}
         \centering
         \includegraphics[width=\textwidth]{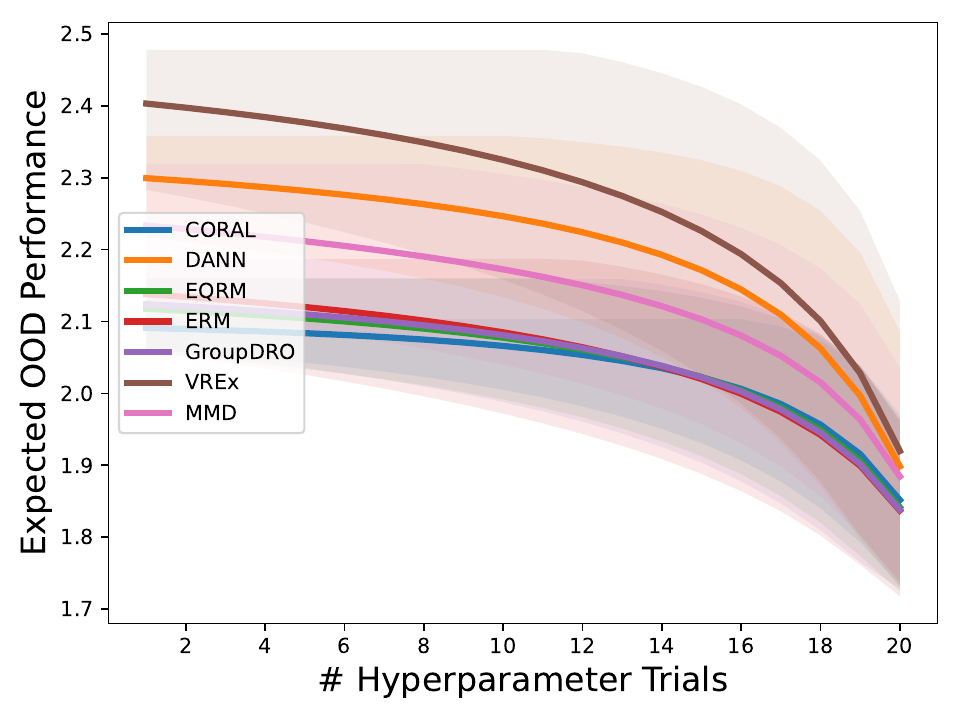}
         \caption{\DWeatherData}
         % \label{fig:five over x}
     \end{subfigure}
        \caption{Expected out-of-distribution performance of various OOD generalization methods in Wild-Tab benchmark with a varying number of hyperparameter search trials. The first row lists all the methods included in the benchmark, while the second row focuses on the "non-sensitive" methods exclusively.}
        \label{fig:evp}
\end{figure}

\paragraph{Hyperparameters Sensitivity in the OOD Generalization}
Shifting our focus to the sensitivity of hyperparameters in OOD generalization, we engage the well-recognized Expected Validation Performance (EVP) technique from NLP~\cite{showyourwork}. Specifically, we use the MAE on \DGTestOut partition within the average-out-domain validation model selection as an EVP measure. The resultant Expected OOD Performance for all assessed methods is visually depicted in \autoref{fig:evp}. While there isn't a universally best method for all scenarios, a meticulous analysis indicates that certain methods, particularly IRM, IB-IRM, IB-ERM, and MMD, display a higher degree of sensitivity to hyperparameter adjustments, therefore demanding more careful configuration for securing reliable results.

% \footnote{This has also been adapted in the Offline Reinforcement Learning as Expected Online Performance~\cite{kurenkov2022eop}}

\paragraph{Beyond OOD Generalization in Tabular Regression} For additional analysis on OOD generalization performance in \DWeatherData classification task and comparison with tree-based baselines such as CatBoost~\cite{catboost} and XGBoost~\cite{xgboost} follow \autoref{appendix:extra_exps}.

\clearpage

\section{Comparison with Existing Benchmarks}

The machine learning research community has a rich history of addressing distribution shifts across a wide array of tasks, evidenced by a myriad of datasets and benchmarks \cite{hand2006classifier,quinonero2009dataset}. However, most of these tasks are based on text or image data such as part-of-speech tagging \cite{marcus93treebank}, sentiment analysis \cite{blitzer2007biographies, fang2014domain}, land cover classification \cite{bruzzone2009domain}, and object recognition and detection \cite{saenko2010adapting}.

Notably, larger-scale, real-world data with distribution shift benchmarks emerged with the advent of datasets like ImageNet-A~\cite{hendrycks2021natural}, ImageNet-C~\cite{hendrycks2019benchmarking}, and CIFAR-10.1~\cite{recht2018cifar}. Following these were the benchmarks with synthetic shifts such as PACS~\cite{li2017deeper}, DomainNet~\cite{peng2019moment}, and others \cite{fang2013unbiased,he2021towards,hendrycks2021many,liang2022metashift,sagawa2019distributionally,santurkar2020breeds,venkateswara2017deep}. While they have undoubtedly contributed to algorithmic advancement, it remains challenging to ensure real-world robustness due to the inherent limitations of synthetic shifts.

Recently, more focus has been placed on real-world distribution shifts. In particular, WILDS~\cite{koh2020wilds,koh2021wilds} and SHIFTS~\cite{malinin2022shifts,malinin2021shifts}, have curated datasets spanning diverse applications and introduced out-of-distribution settings to facilitate robustness testing. However, their focus is primarily on deep learning methods generalization or on the tabular data domain and thus is somewhat orthogonal.

Contrarily, our work addresses a critical gap in the current benchmarks by focusing on out-of-distribution deep learning generalization, specifically in the context of tabular data. Our proposed benchmark, Wild-Tab, is designed to provide a comprehensive testbed for out-of-distribution generalization in tabular regression, thus offering the research community a dedicated and relevant resource for evaluating and developing novel methods in this critical domain.

\section{Conclusion and Discussion}
\label{section:conclusion}

In this paper, we introduce Wild-Tab, a novel benchmark devised to examine the effect of in-the-wild distribution shifts on out-of-distribution methods in tabular regression. 
Notably, we emphasize the exploitation of tabular data, a form of data underrepresented in extant robustness methodologies and benchmarks.
In our study, we conducted a comprehensive evaluation of 10 distinct OOD methods across 3 large-scale datasets. 
The empirical findings highlight a marked discrepancy between in-distribution and out-of-distribution performance across all tasks, an outcome attributed to the underlying distribution shift. 
Our observations showed that none of the current OOD generalization methods is invariably more resilient to these shifts than Empirical Risk Minimization, suggesting that there is still room for improvement for OOD methods to handle such shifts more effectively.

By uncovering the limitations of existing OOD generalization approaches in the face of distributional shifts, we present an opportunity for advancements in this field. Our results highlight the necessity of developing more robust methods that can effectively cope with distribution shifts and thus have the potential to enhance real-world applications where data change is a norm.

Despite the available findings, our research is not without its limitations. We primarily focused on tabular data, which might not represent all types of data. The ability of our findings to generalize to other data types remains a subject for further exploration. Furthermore, the lack of a consistently superior approach to ERM might be attributed to the selection of methods benchmarked, suggesting that there could be other, unexplored methods that may perform better. These limitations do not invalidate our conclusions but rather qualify them, serving as potential starting points for future research.

In conclusion, our work elucidates the under-studied impact of distribution shifts on OOD performance in tabular regression tasks. 
We have offered the Wild-Tab benchmark as a tool for future research, shedding light on the limitations of current OOD generalization approaches and highlighting the need for more resilient methodologies.
It is our aspiration that these novel methodologies will prove to be trustworthy and effective when implemented in real-world environments.

% \section*{References}
\bibliographystyle{plain}
\bibliography{references.bib}

\newpage
\clearpage
\appendix

% \clearpage
\section{Datasets}
\label{appendix:datasets}

\begin{figure}[h]
     \centering
     \begin{subfigure}[b]{0.96\textwidth}
         \centering
         \includegraphics[width=0.48\textwidth]{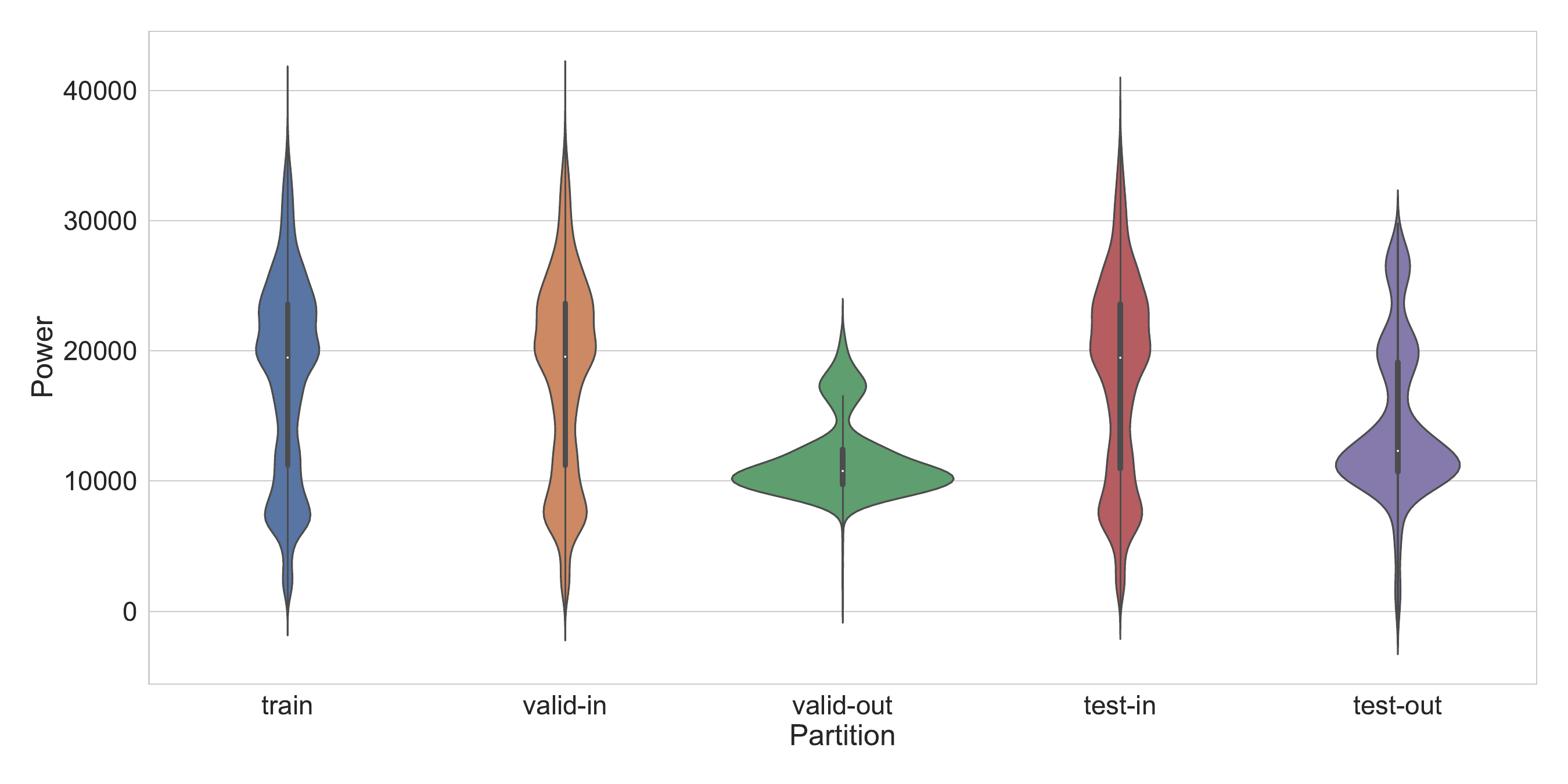}
         \includegraphics[width=0.48\textwidth]{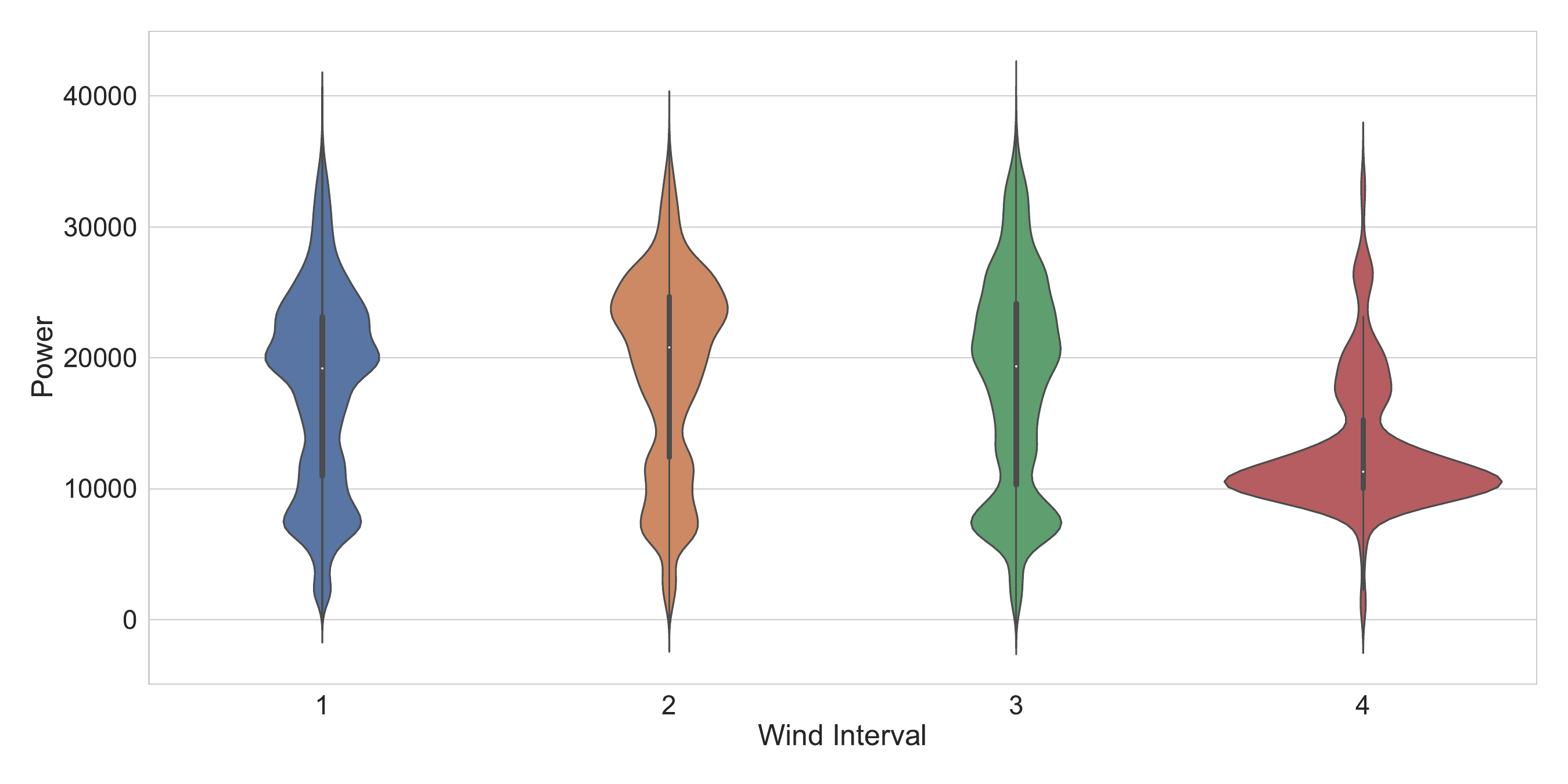}
         \caption{\DVpowerSynt}
     \end{subfigure}
     \hfill
     \begin{subfigure}[b]{0.96\textwidth}
         \centering
         \includegraphics[width=0.48\textwidth]{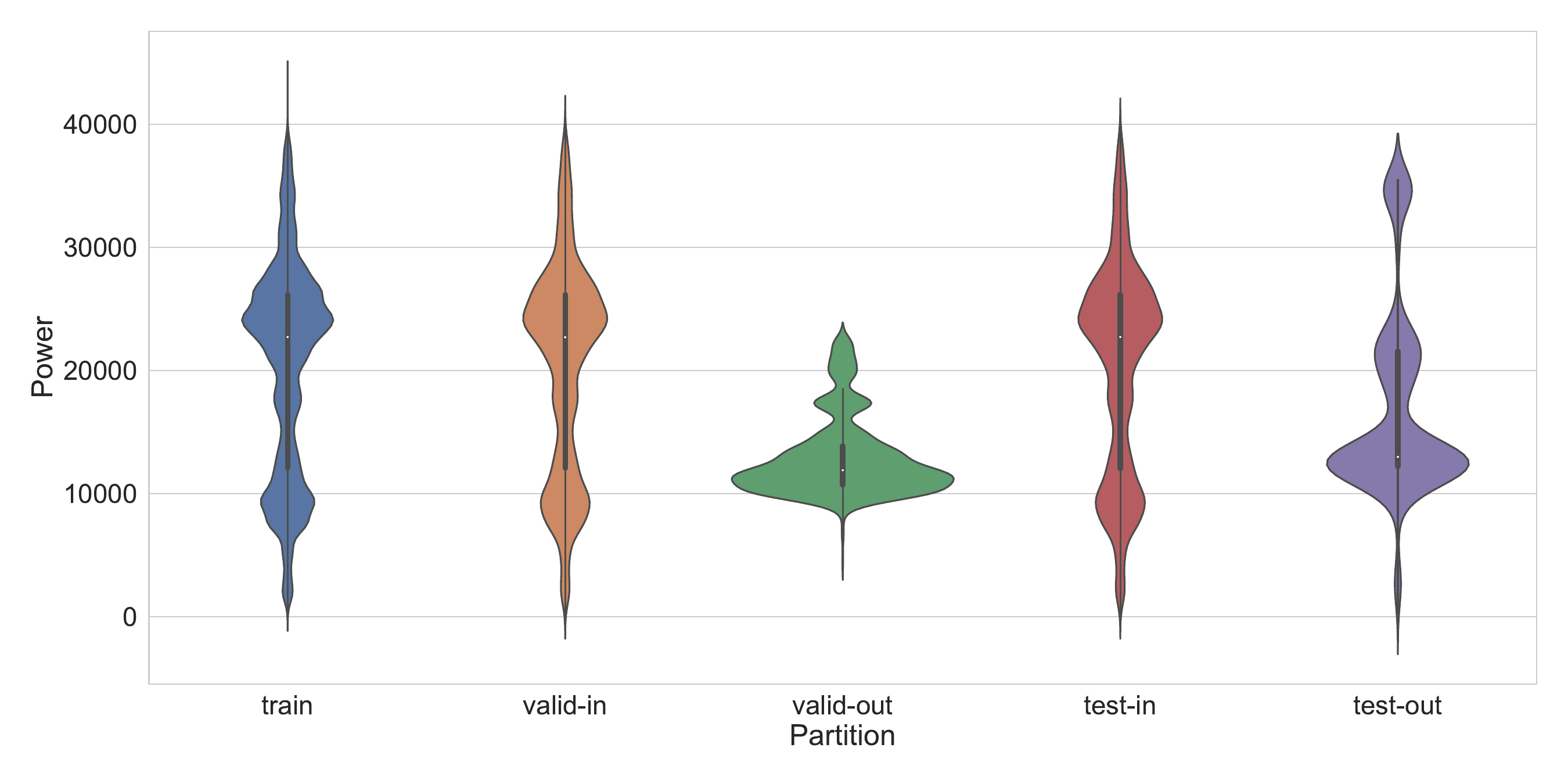}
         \includegraphics[width=0.48\textwidth]{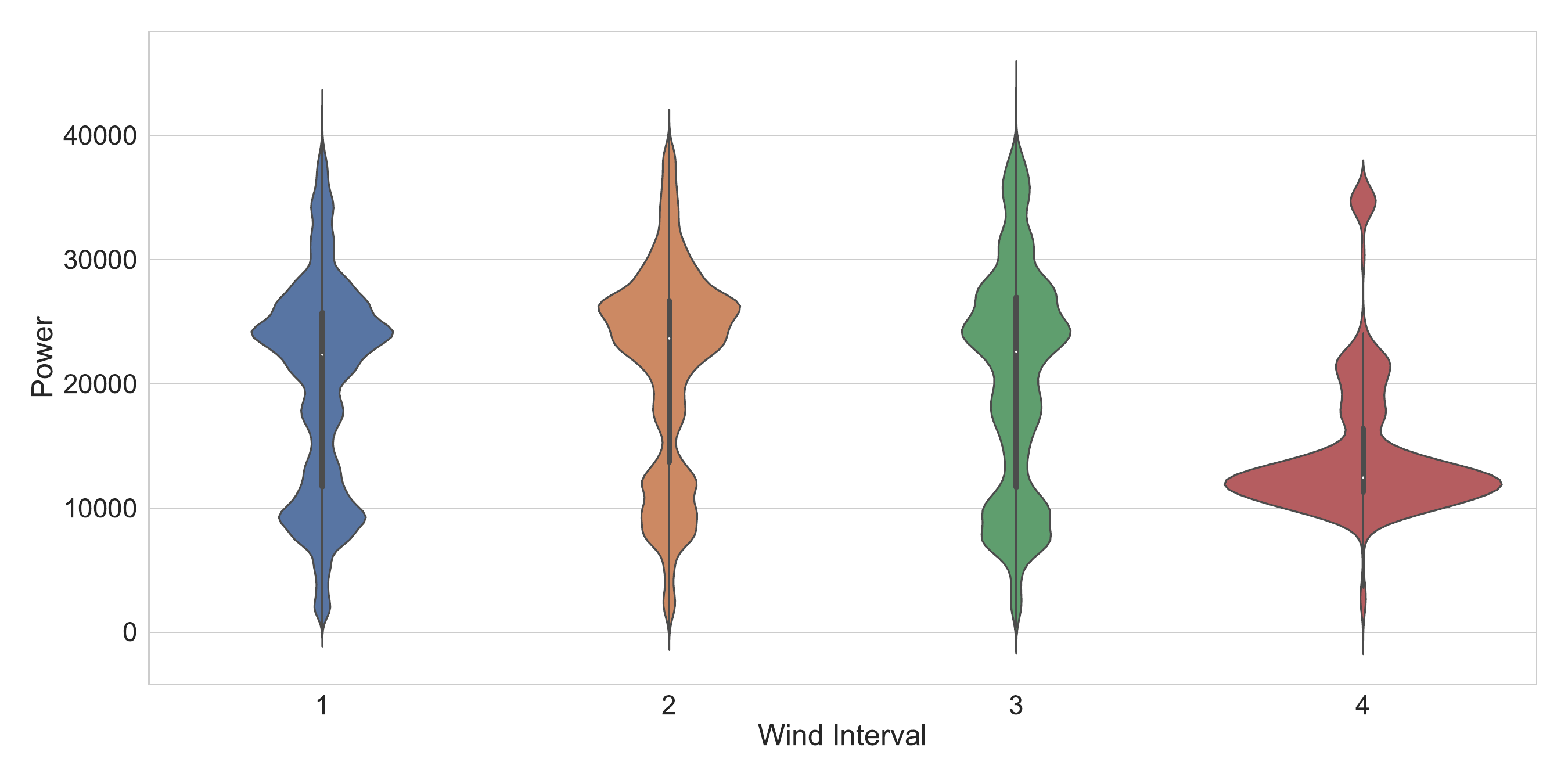}
         \caption{\DVpowerReal}
     \end{subfigure}
     \hfill
     \begin{subfigure}[b]{0.96\textwidth}
         \centering
         \includegraphics[width=0.3\textwidth]{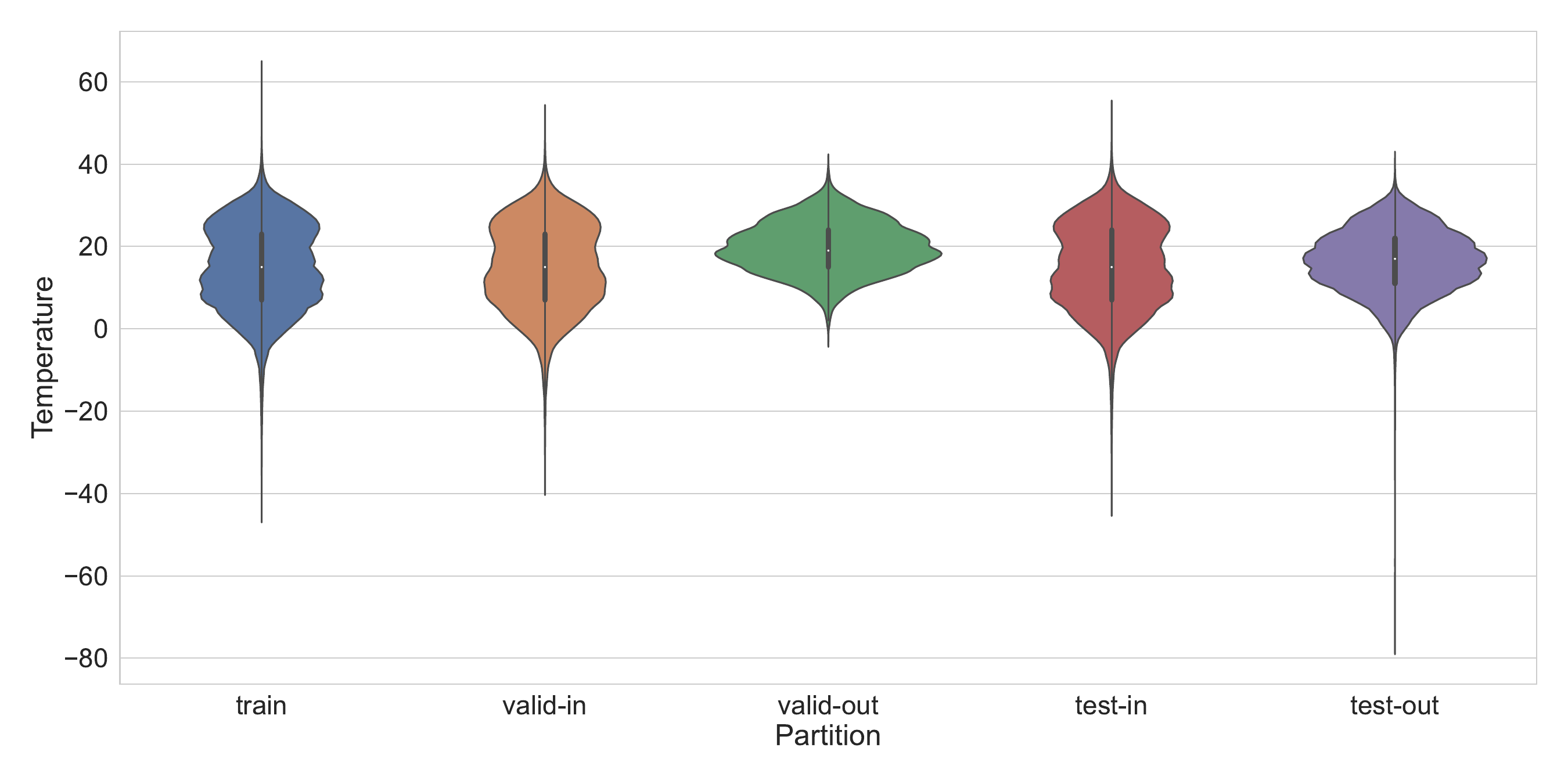}
         \includegraphics[width=0.3\textwidth]{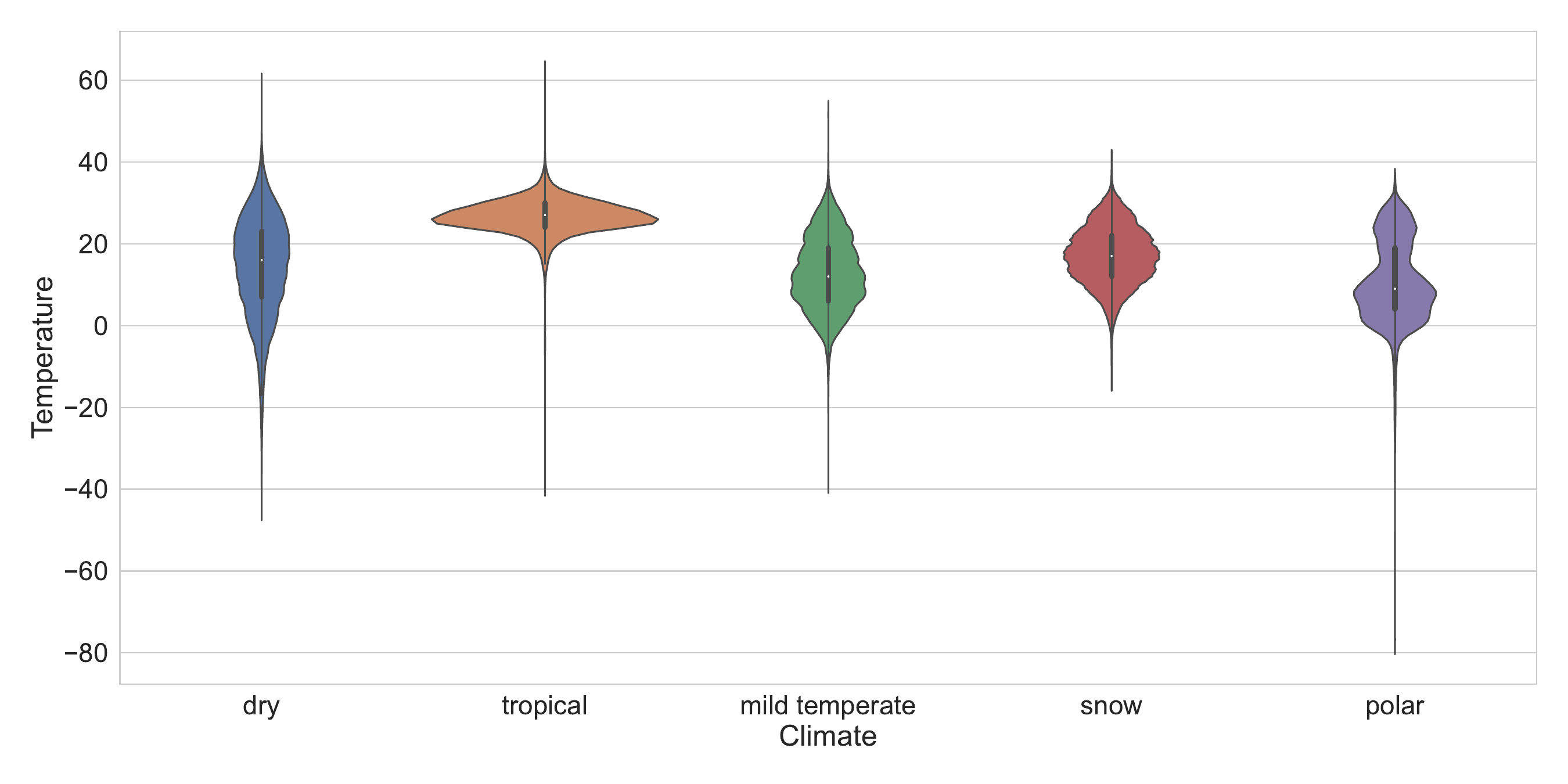}
         \includegraphics[width=0.3\textwidth]{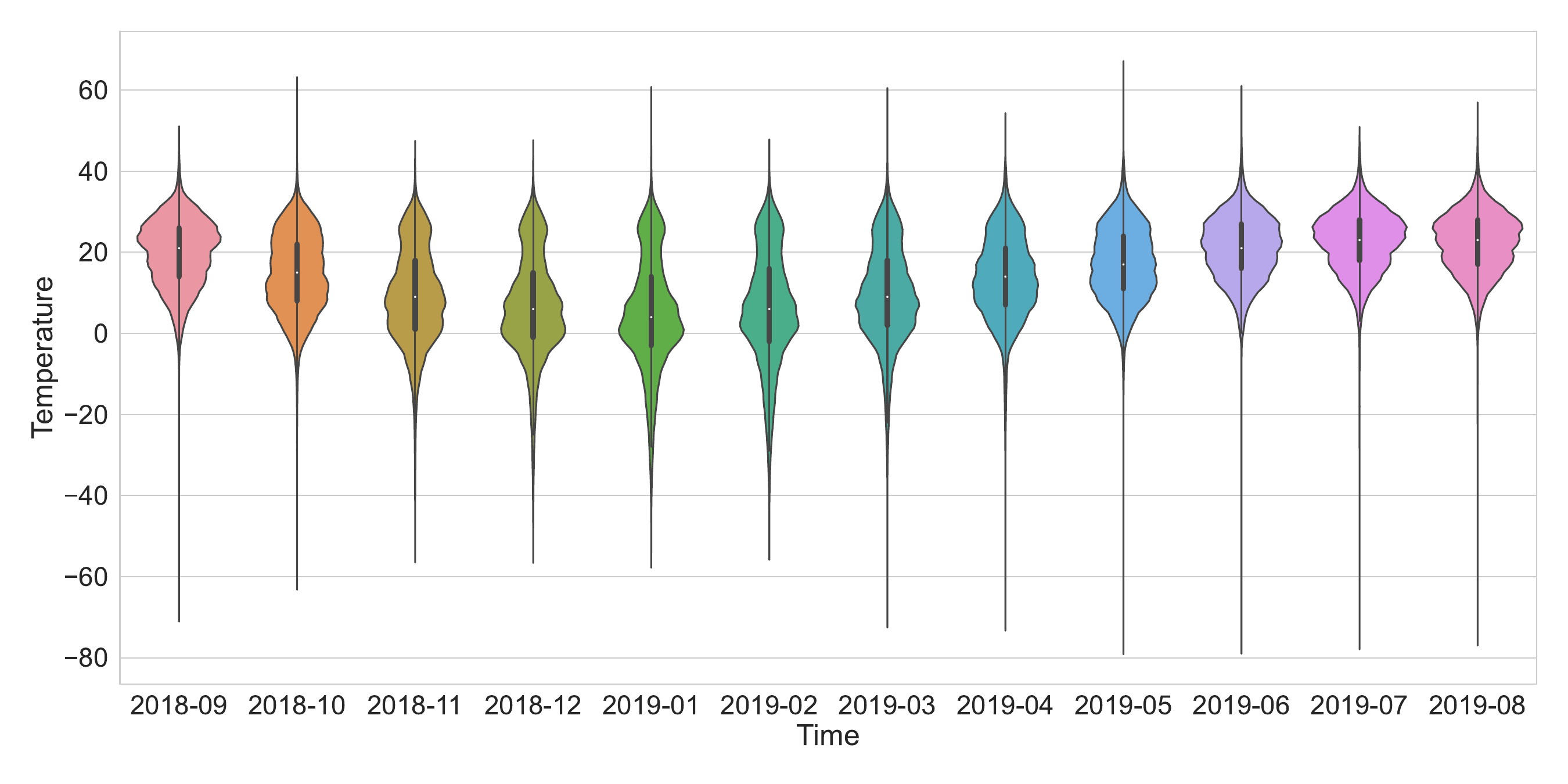}
         \caption{\DWeatherData}
     \end{subfigure}
    \caption{Targets distributions of benchmarked datasets.}
    \label{fig:datasets_eda}
\end{figure}

In our experiments, we utilize datasets introduced in previous works \cite{malinin2022shifts, malinin2021shifts}. These datasets include the Shifts vessel power estimation dataset and the Shifts Weather Prediction dataset~\footnote{\url{https://github.com/Shifts-Project/shifts}}. For the Weather Prediction dataset, we use a publicly available canonical partition. However, as the test part of the Shifts vessel power estimation dataset is not publicly accessible, we follow the partitioning guidelines outlined in \cite{malinin2022shifts} to create new train and validation partitions. Specifically, we use the $awind\_vcomp\_provider$ feature for data partitioning, as shown in \autoref{table:ds_wind_intervals}. More detailed descriptions of these datasets can be found in the aforementioned works \cite{malinin2022shifts, malinin2021shifts}, and the resulting splits are available in the codebase. Additionally, the target distributions for the datasets we employ are presented in \autoref{fig:datasets_eda}.

\begin{table}[htp]
\caption{Wind intervals considered for data partitioning. Beaufort ranges are defined approximately.}
\centering
    \begin{small}
        \begin{tabular}{ccc}
        \toprule
        Wind interval & Range (kn) & Range in Beaufort \\
        \midrule
        1  & [0, 9) & Up to ~3 \\
        2 & [9, 14) & 3-4 \\
        3 & [14, 19) & 4-5 \\
        4 & $\geq19$ & $\geq5$ \\
        \bottomrule
        \end{tabular}
    \end{small}
\label{table:ds_wind_intervals}
\end{table}

\clearpage
\section{Implementation Details}
\label{appendix:imp_details}

During our experiment, we maintained adherence to the dataset-related hyperparameter recommendations detailed in \cite{malinin2022shifts, malinin2021shifts}, the MLP-related guidelines presented in \cite{gorishniy2022embeddings}, and the suggested hyperparameter space for OOD generalization methods in \cite{gulrajani2020search}. We conducted a total of 4200 runs using an NVIDIA V100 GPU.

\subsection{Dataset-specific Hyperparameters}

The experimental setups for the \DVpowerSynt and \DVpowerSynt datasets involve a batch size of 128 and an MLP width of 64. In the case of the \DWeatherData dataset, we employ a batch size of 1024 and an MLP width of 512. Across all datasets, our experiments are conducted over 100 iterations.

\subsection{Model-specific Hyperparameters}

\begin{table}[h]
\centering
\caption{MLP hyperparameter space.}
\label{tab:A-mlp-space}
% \vspace{1em}
{\renewcommand{\arraystretch}{1.2}
\resizebox{0.55\textwidth}{!}{
\begin{tabular}{lll}
    \toprule
    \textbf{Parameter} & \textbf{Default value} &\textbf{Random distribution} \\
    \midrule
    \# Layers & 4 &$\mathrm{UniformInt}[1,8]$ \\
    Dropout & 0 & $\mathrm{Uniform}[0, 0.5]$ \\
    Learning rate & $1e\text{-}4$ & $\mathrm{LogUniform}[5e\text{-}5, 0.005]$ \\
    Weight decay & 0 & $\mathrm{LogUniform}[1e\text{-}6, 1e\text{-}3]$ \\
    % \midrule
    % \# Iterations & 100 \\
    \bottomrule
\end{tabular}}
}
\end{table}

\newcommand{\PLE}{\texttt{PLE}}
\paragraph{Numerical Feature Embeddings Hyperparameters}
The distribution for the output dimensions of linear layers is $\mathrm{UniformInt}[1, 128]$.
For models based on the MLP-PLR, the distribution for $\sigma$ is $\mathrm{LogUniform}[1e\text{-}2, 1e2]$ (it is the same for all features) and the distribution for $k$ is $\mathrm{UniformInt}[1,128]$.
For the target-aware (tree-based) \PLE~(MLP-T-LR), the distribution for the number of leaves is $\mathrm{UniformInt}[2,256]$, the distribution for the minimum number of items per leaf is $\mathrm{UniformInt}[1,128]$, and the distribution for the minimum information gain required for making a split is $\mathrm{LogUniform}[1e\text{-}9, 0.01]$.

\subsection{Objective-specific Hyperparameters}
\begin{table}[h]
    \caption{Objectives hyperparameter space.} 
    \begin{center}
    {
    \resizebox{0.7\textwidth}{!}{
    \begin{tabular}{llll}
        \toprule
        \textbf{Objective} & \textbf{Parameter} & \textbf{Default value} & \textbf{Random distribution}\\
        \midrule
        CORAL                           & penalty weight & 1 & $10^{\text{Uniform}(-1, 1)}$\\
        \midrule
        \multirow{5}{*}{DANN} & lambda                 & 1.0    & $10^{\text{Uniform}(-2, 2)}$\\
         % & discriminator weight decay & 0    & $10^{\text{Uniform}(-6, -2)}$\\
         & discriminator steps        & 1    & $2^{\text{Uniform}(0, 3)}$\\
         & gradient penalty           & 0    & $10^{\text{Uniform}(-2, 1)}$\\
         & adam $\beta_1$             & 0.5    & $\text{RandomChoice}([0, 0.5])$\\
        \midrule
        \multirow{2}{*}{EQRM}          & quantile  & 0.75    & $\text{Uniform}(0.5, 0.99)$\\
                                      & burnin iterations & 500 & $10^{\text{Uniform}(2.5, 3.5)}$\\
        \midrule
        GroupDRO                           & eta   & 0.01 & $10^{\text{Uniform}(-3, 1)}$\\
        \midrule
        \multirow{2}{*}{IB-ERM} & IB penalty weight  & 100    & $10^{\text{Uniform}(-1, 5)}$\\
                                & IB iterations of penalty annealing & 500 & $10^{\text{Uniform}(0, 4)}$\\
        \midrule
        \multirow{2}{*}{IB-IRM} & IB penalty weight  & 100    & $10^{\text{Uniform}(-1, 5)}$\\
                                & IB iterations of penalty annealing & 500 & $10^{\text{Uniform}(0, 4)}$\\
                                & IRM penalty weight  & 100    & $10^{\text{Uniform}(-1, 5)}$\\
                                & IRM iterations of penalty annealing & 500 & $10^{\text{Uniform}(0, 4)}$\\
        \midrule                 
        \multirow{2}{*}{IRM}          & penalty weight  & 100    & $10^{\text{Uniform}(-1, 5)}$\\
                                      & iterations of penalty annealing & 500 & $10^{\text{Uniform}(0, 4)}$\\
        \midrule
        MMD                           & penalty weight & 1 & $10^{\text{Uniform}(-1, 1)}$\\
        \midrule
        \multirow{2}{*}{VREx}          & penalty weight  & 10    & $10^{\text{Uniform}(-1, 5)}$\\
                                      & iterations of penalty annealing & 500 & $10^{\text{Uniform}(0, 4)}$\\
        \bottomrule
    \end{tabular}
    }
    }
    \end{center}
    \label{table:hyperparameters}
\end{table}

\clearpage
\section{Additional Experimental Results}
\label{appendix:extra_exps}

\paragraph{Analysis on Tree-Based Approaches Versus Neural Networks} To further elucidate the landscape of OOD generalization, we juxtaposed the ERM approach with prominent tree-based contenders, namely catboost and xgboost, as delineated in \autoref{table:BenchmarkResultsGB}. The empirical findings, intriguingly, indicate that tree-based methodologies, although often preferred for tabular data, don't exhibit superior performance on the datasets examined in this study.

% A plausible hypothesis for this phenomenon hinges on the inherent limitations of tree-based techniques in extrapolative capabilities. This capability becomes pivotal when navigating a domain generalization context ($\Domtrain \cap \Domtest=\emptyset$) - a hallmark feature of the Wild-Tab benchmark. 
A plausible reason underscoring this discrepancy might revolve around the inherent inability of tree-based methods to extrapolate beyond their training domain. Such extrapolation is pivotal for domain generalization ($\Domtrain \cap \Domtest=\emptyset$) -- the Wild-Tab benchmark case.

\begin{table}[htbp!]
\centering
\caption{ERM comparison with tree-based alternatives on Wild-Tab.}
\label{table:BenchmarkResultsGB}
\resizebox{0.9\textwidth}{!}{
\begin{tabular}{l|l|ll|ll|ll}
\toprule
\multirow{2}*{Data} & \multirow{2}*{Objective} & \multicolumn{2}{c|}{\DVpowerSynt} & \multicolumn{2}{c|}{\DVpowerReal} & \multicolumn{2}{c}{\DWeatherData} \\
& & In & Out & In & Out & In & Out \\
\midrule

\multirow{3}*{Valid} & ERM & $797_{\pm1}$ & $751_{\pm22}$ & $624_{\pm27}$ & $901_{\pm49}$ & $1.295_{\pm0.005}$ & $1.674_{\pm0.017}$ \\
& catboost & $911_{\pm0}$ & $734_{\pm0}$ & $875_{\pm0}$ & $813_{\pm0}$ & $1.557_{\pm0.0}$ & $1.682_{\pm0.0}$ \\
& xgboost & $910_{\pm0}$ & $741_{\pm0}$ & $1002_{\pm0}$ & $934_{\pm0}$ & $1.483_{\pm0.0}$ & $1.664_{\pm0.0}$ \\
\midrule
\multirow{3}*{Test} & ERM & $798_{\pm2}$ & $1022_{\pm12}$ & $619_{\pm27}$ & $2253_{\pm376}$ & $1.291_{\pm0.003}$ & $1.803_{\pm0.007}$ \\
& catboost & $912_{\pm0}$ & $1168_{\pm0}$ & $869_{\pm0}$ & $1719_{\pm0}$ & $1.557_{\pm0.0}$ & $1.87_{\pm0.0}$ \\
& xgboost & $915_{\pm0}$ & $1461_{\pm0}$ & $998_{\pm0}$ & $2035_{\pm0}$ & $1.482_{\pm0.0}$ & $1.828_{\pm0.0}$ \\
\midrule
\end{tabular}
}
\end{table}

\begin{wraptable}{L}{6cm}
% \begin{table}[htbp!]
\centering
\caption{OOD generalization methods accuracy on \DWeatherData dataset with average-out-domain validation.}
\label{table:BenchmarkResultsCLF}
\resizebox{0.4\textwidth}{!}{
\begin{tabular}{l|l|ll|ll|ll}
\toprule
Data & Objective & In & Out \\
% \multirow{2}*{Data} & \multirow{2}*{Objective} & \multicolumn{2}{c|}{\DVpowerSynt} & \multicolumn{2}{c|}{\DVpowerReal} & \multicolumn{2}{c}{\DWeatherData} \\
% & & In & Out & In & Out & In & Out \\
\midrule

\multirow{10}*{Valid} & CORAL & $60.9_{\pm0.2}$ & $51.8_{\pm0.3}$ \\
& DANN & $61.8_{\pm0.6}$ & $51.7_{\pm0.2}$ \\
& EQRM & $62.5_{\pm0.4}$ & $51.7_{\pm0.2}$ \\
& ERM & $61.1_{\pm0.1}$ & $51.9_{\pm0.3}$ \\
& GroupDRO & $61.0_{\pm0.1}$ & $51.9_{\pm0.2}$ \\
& IB\_ERM & $62.2_{\pm1.6}$ & $51.7_{\pm0.1}$ \\
& IB\_IRM & $60.5_{\pm0.0}$ & $51.6_{\pm0.2}$ \\
& IRM & $61.7_{\pm0.1}$ & $51.7_{\pm0.4}$ \\
& MMD & $60.8_{\pm0.1}$ & $51.9_{\pm0.3}$ \\
& VREx & $61.1_{\pm0.1}$ & $51.6_{\pm0.2}$ \\
\midrule
\multirow{10}*{Test} & CORAL & $60.6_{\pm0.1}$ & $48.4_{\pm0.1}$ \\
& DANN & $61.6_{\pm0.6}$ & $48.1_{\pm0.1}$ \\
& EQRM & $62.3_{\pm0.3}$ & $48.3_{\pm0.3}$ \\
& ERM & $60.8_{\pm0.1}$ & $48.6_{\pm0.3}$ \\
& GroupDRO & $60.7_{\pm0.0}$ & $48.6_{\pm0.2}$ \\
& IB\_ERM & $61.9_{\pm1.5}$ & $48.2_{\pm0.1}$ \\
& IB\_IRM & $60.3_{\pm0.0}$ & $48.1_{\pm0.1}$ \\
& IRM & $61.5_{\pm0.1}$ & $48.3_{\pm0.2}$ \\
& MMD & $60.6_{\pm0.1}$ & $48.5_{\pm0.2}$ \\
& VREx & $60.8_{\pm0.1}$ & $48.5_{\pm0.2}$ \\
\bottomrule
\end{tabular}
}
% \end{table}
\end{wraptable}

\paragraph{OOD Generalization in Tabular Classification} In broadening our examination scope, we conducted further experiments on the large-scale \DWeatherData dataset classification task, the details of which are outlined in \autoref{table:BenchmarkResultsCLF}. Our findings, albeit exhibiting slight deviations in terms of model performance and rankings, echoed previously established patterns. Specifically, we identified: (1) a prominent generalization gap, (2) an unpredictable performance on test sets, and (3) slight advancements over the Empirical Risk Minimization (ERM) baseline. These trends once again underscore the challenges and nuances of OOD generalization in tabular contexts.

\paragraph{Evaluating with Varied Training Data Volumes} In our investigation, we assessed the efficacy of all baselines in the context of a reduced number of training examples. Specifically, within the benchmark framework, we employed subsets amounting to 33\% and 66\% of the total training data, as opposed to the full dataset referenced in our initial findings (\autoref{table:BenchmarkResults} in the main paper). The outcomes of these evaluations are detailed in \autoref{table:BenchmarkResultsFRAC}.

Our findings indicate that while the principal conclusions from our main benchmark remain consistent across varying volumes of training data, an intriguing pattern emerges. Although the test in-distribution error demonstrates improvement as the volume of training data expands, the test out-of-distribution error remains relatively consistent. Thus once again underscoring the distinct challenges posed by OOD generalization. Moreover, such a finding accentuates the imperative need for innovative methods, ones that can more effectively leverage expansive datasets to achieve superior generalization within OOD contexts.
% It accentuates the imperative need for innovative methods, ones that can more effectively leverage expansive datasets to achieve superior generalization within in-distribution contexts.

\begin{table}[htbp!]
\centering
\caption{
OOD generalization performance of all baselines when reducing the amount of training data. We randomly allocate 33\% or 66\% of the data for training rather than 100\% in our original benchmark.
}
\label{table:BenchmarkResultsFRAC}
\resizebox{1.0\textwidth}{!}{
\begin{tabular}{ll|l|ll|ll|ll}
\toprule
\multirow{2}*{$\%$ Train} & \multirow{2}*{Data} & \multirow{2}*{Objective} & \multicolumn{2}{c|}{\DVpowerSynt} & \multicolumn{2}{c|}{\DVpowerReal} & \multicolumn{2}{c}{\DWeatherData} \\
& & & In & Out & In & Out & In & Out \\
\midrule

\multirow{20}*{$33\%$} & \multirow{10}*{Valid} & CORAL & $921_{\pm38}$ & $700_{\pm4}$ & $1045_{\pm46}$ & $650_{\pm62}$ & $1.409_{\pm0.002}$ & $1.626_{\pm0.013}$ \\
& & DANN & $895_{\pm20}$ & $703_{\pm0}$ & $997_{\pm37}$ & $674_{\pm22}$ & $1.433_{\pm0.012}$ & $1.616_{\pm0.009}$ \\
& & EQRM & $920_{\pm48}$ & $697_{\pm5}$ & $1186_{\pm85}$ & $670_{\pm34}$ & $1.415_{\pm0.01}$ & $1.629_{\pm0.009}$ \\
& & ERM & $934_{\pm43}$ & $700_{\pm3}$ & $1006_{\pm48}$ & $650_{\pm44}$ & $1.437_{\pm0.009}$ & $1.615_{\pm0.001}$ \\
& & GroupDRO & $913_{\pm48}$ & $703_{\pm2}$ & $1115_{\pm26}$ & $667_{\pm20}$ & $1.419_{\pm0.012}$ & $1.623_{\pm0.004}$ \\
& & IB\_ERM & $926_{\pm56}$ & $702_{\pm3}$ & $1153_{\pm91}$ & $661_{\pm26}$ & $1.475_{\pm0.005}$ & $1.647_{\pm0.006}$ \\
& & IB\_IRM & $942_{\pm57}$ & $717_{\pm17}$ & $1016_{\pm32}$ & $640_{\pm18}$ & $1.488_{\pm0.017}$ & $1.658_{\pm0.002}$ \\
& & IRM & $1046_{\pm14}$ & $698_{\pm3}$ & $1008_{\pm23}$ & $659_{\pm45}$ & $1.402_{\pm0.003}$ & $1.62_{\pm0.01}$ \\
& & MMD & $953_{\pm12}$ & $698_{\pm6}$ & $1085_{\pm83}$ & $650_{\pm13}$ & $1.414_{\pm0.006}$ & $1.629_{\pm0.006}$ \\
& & VREx & $977_{\pm69}$ & $702_{\pm6}$ & $1018_{\pm35}$ & $661_{\pm46}$ & $1.423_{\pm0.022}$ & $1.63_{\pm0.005}$ \\
\cmidrule{2-9}
 & \multirow{10}*{Test} & CORAL & $928_{\pm35}$ & $928_{\pm13}$ & $1028_{\pm45}$ & $1762_{\pm116}$ & $1.409_{\pm0.003}$ & $1.755_{\pm0.005}$ \\
& & DANN & $901_{\pm25}$ & $947_{\pm30}$ & $983_{\pm20}$ & $2011_{\pm86}$ & $1.431_{\pm0.012}$ & $1.763_{\pm0.012}$ \\
& & EQRM & $927_{\pm46}$ & $1061_{\pm39}$ & $1177_{\pm87}$ & $1587_{\pm173}$ & $1.414_{\pm0.009}$ & $1.76_{\pm0.006}$ \\
& & ERM & $937_{\pm43}$ & $969_{\pm40}$ & $993_{\pm58}$ & $1978_{\pm94}$ & $1.438_{\pm0.007}$ & $1.755_{\pm0.006}$ \\
& & GroupDRO & $915_{\pm47}$ & $948_{\pm25}$ & $1104_{\pm17}$ & $1615_{\pm92}$ & $1.419_{\pm0.011}$ & $1.752_{\pm0.012}$ \\
& & IB\_ERM & $929_{\pm62}$ & $1015_{\pm75}$ & $1142_{\pm100}$ & $1921_{\pm396}$ & $1.475_{\pm0.005}$ & $1.782_{\pm0.014}$ \\
& & IB\_IRM & $948_{\pm50}$ & $1150_{\pm15}$ & $997_{\pm34}$ & $1802_{\pm114}$ & $1.488_{\pm0.019}$ & $1.774_{\pm0.016}$ \\
& & IRM & $1058_{\pm18}$ & $1032_{\pm35}$ & $994_{\pm32}$ & $1737_{\pm161}$ & $1.404_{\pm0.002}$ & $1.766_{\pm0.005}$ \\
& & MMD & $958_{\pm3}$ & $947_{\pm3}$ & $1066_{\pm78}$ & $1795_{\pm49}$ & $1.412_{\pm0.004}$ & $1.764_{\pm0.012}$ \\
& & VREx & $987_{\pm67}$ & $964_{\pm30}$ & $1000_{\pm30}$ & $2032_{\pm208}$ & $1.42_{\pm0.022}$ & $1.764_{\pm0.012}$ \\
\midrule
\multirow{20}*{$66\%$} & \multirow{10}*{Valid} & CORAL & $969_{\pm22}$ & $699_{\pm3}$ & $1014_{\pm35}$ & $636_{\pm6}$ & $1.45_{\pm0.017}$ & $1.61_{\pm0.022}$ \\
& & DANN & $921_{\pm30}$ & $697_{\pm4}$ & $997_{\pm35}$ & $676_{\pm58}$ & $1.407_{\pm0.022}$ & $1.605_{\pm0.014}$ \\
& & EQRM & $1005_{\pm18}$ & $706_{\pm20}$ & $1085_{\pm15}$ & $602_{\pm17}$ & $1.405_{\pm0.007}$ & $1.599_{\pm0.005}$ \\
& & ERM & $935_{\pm7}$ & $698_{\pm10}$ & $1057_{\pm59}$ & $612_{\pm12}$ & $1.394_{\pm0.025}$ & $1.609_{\pm0.007}$ \\
& & GroupDRO & $900_{\pm12}$ & $699_{\pm3}$ & $1056_{\pm72}$ & $654_{\pm36}$ & $1.378_{\pm0.033}$ & $1.601_{\pm0.006}$ \\
& & IB\_ERM & $1003_{\pm17}$ & $707_{\pm10}$ & $1193_{\pm59}$ & $638_{\pm20}$ & $1.415_{\pm0.028}$ & $1.636_{\pm0.01}$ \\
& & IB\_IRM & $904_{\pm5}$ & $701_{\pm10}$ & $1073_{\pm114}$ & $647_{\pm55}$ & $1.522_{\pm0.028}$ & $1.638_{\pm0.009}$ \\
& & IRM & $925_{\pm33}$ & $701_{\pm2}$ & $969_{\pm25}$ & $625_{\pm30}$ & $1.403_{\pm0.007}$ & $1.607_{\pm0.006}$ \\
& & MMD & $1038_{\pm189}$ & $699_{\pm6}$ & $1102_{\pm71}$ & $598_{\pm46}$ & $1.386_{\pm0.01}$ & $1.612_{\pm0.003}$ \\
& & VREx & $953_{\pm62}$ & $700_{\pm5}$ & $1200_{\pm61}$ & $647_{\pm39}$ & $1.396_{\pm0.019}$ & $1.615_{\pm0.014}$ \\
\cmidrule{2-9}
 & \multirow{10}*{Test} & CORAL & $976_{\pm22}$ & $1050_{\pm42}$ & $1009_{\pm26}$ & $1806_{\pm109}$ & $1.45_{\pm0.015}$ & $1.765_{\pm0.025}$ \\
& & DANN & $926_{\pm31}$ & $965_{\pm67}$ & $988_{\pm27}$ & $1762_{\pm120}$ & $1.406_{\pm0.024}$ & $1.74_{\pm0.017}$ \\
& & EQRM & $1008_{\pm20}$ & $1082_{\pm76}$ & $1077_{\pm18}$ & $1681_{\pm8}$ & $1.405_{\pm0.006}$ & $1.734_{\pm0.009}$ \\
& & ERM & $939_{\pm8}$ & $1043_{\pm34}$ & $1046_{\pm56}$ & $1760_{\pm135}$ & $1.393_{\pm0.027}$ & $1.729_{\pm0.004}$ \\
& & GroupDRO & $905_{\pm14}$ & $950_{\pm30}$ & $1047_{\pm63}$ & $1811_{\pm108}$ & $1.379_{\pm0.033}$ & $1.736_{\pm0.014}$ \\
& & IB\_ERM & $1004_{\pm18}$ & $1071_{\pm92}$ & $1180_{\pm59}$ & $2025_{\pm184}$ & $1.412_{\pm0.027}$ & $1.768_{\pm0.013}$ \\
& & IB\_IRM & $908_{\pm1}$ & $1047_{\pm109}$ & $1063_{\pm113}$ & $1872_{\pm124}$ & $1.519_{\pm0.029}$ & $1.806_{\pm0.043}$ \\
& & IRM & $931_{\pm33}$ & $965_{\pm9}$ & $959_{\pm22}$ & $1912_{\pm345}$ & $1.401_{\pm0.008}$ & $1.725_{\pm0.004}$ \\
& & MMD & $1044_{\pm194}$ & $948_{\pm19}$ & $1092_{\pm74}$ & $1839_{\pm135}$ & $1.381_{\pm0.009}$ & $1.741_{\pm0.011}$ \\
& & VREx & $956_{\pm63}$ & $944_{\pm5}$ & $1189_{\pm64}$ & $1880_{\pm79}$ & $1.396_{\pm0.019}$ & $1.738_{\pm0.015}$ \\
\midrule
\multirow{20}*{$100\%$} & \multirow{10}*{Valid} & CORAL & $862_{\pm9}$ & $700_{\pm5}$ & $1008_{\pm35}$ & $618_{\pm33}$ & $1.35_{\pm0.018}$ & $1.604_{\pm0.013}$ \\
& & DANN & $988_{\pm77}$ & $704_{\pm7}$ & $973_{\pm5}$ & $632_{\pm21}$ & $1.474_{\pm0.011}$ & $1.604_{\pm0.009}$ \\
& & EQRM & $878_{\pm14}$ & $705_{\pm9}$ & $937_{\pm33}$ & $620_{\pm29}$ & $1.457_{\pm0.038}$ & $1.61_{\pm0.005}$ \\
& & ERM & $872_{\pm11}$ & $705_{\pm6}$ & $1002_{\pm36}$ & $628_{\pm10}$ & $1.357_{\pm0.026}$ & $1.603_{\pm0.008}$ \\
& & GroupDRO & $873_{\pm18}$ & $705_{\pm3}$ & $1000_{\pm47}$ & $635_{\pm44}$ & $1.36_{\pm0.03}$ & $1.599_{\pm0.004}$ \\
& & IB\_ERM & $900_{\pm22}$ & $716_{\pm7}$ & $932_{\pm15}$ & $610_{\pm5}$ & $1.397_{\pm0.008}$ & $1.62_{\pm0.007}$ \\
& & IB\_IRM & $860_{\pm3}$ & $714_{\pm4}$ & $919_{\pm16}$ & $632_{\pm53}$ & $1.418_{\pm0.025}$ & $1.623_{\pm0.008}$ \\
& & IRM & $869_{\pm8}$ & $704_{\pm2}$ & $1032_{\pm16}$ & $673_{\pm56}$ & $1.352_{\pm0.048}$ & $1.603_{\pm0.002}$ \\
& & MMD & $865_{\pm14}$ & $700_{\pm4}$ & $979_{\pm22}$ & $639_{\pm18}$ & $1.371_{\pm0.008}$ & $1.607_{\pm0.008}$ \\
& & VREx & $937_{\pm21}$ & $709_{\pm8}$ & $950_{\pm18}$ & $647_{\pm8}$ & $1.355_{\pm0.023}$ & $1.601_{\pm0.001}$ \\
\cmidrule{2-9}
 & \multirow{10}*{Test} & CORAL & $865_{\pm10}$ & $916_{\pm8}$ & $1001_{\pm33}$ & $1679_{\pm184}$ & $1.345_{\pm0.019}$ & $1.741_{\pm0.005}$ \\
& & DANN & $993_{\pm77}$ & $1029_{\pm63}$ & $965_{\pm2}$ & $1714_{\pm78}$ & $1.471_{\pm0.01}$ & $1.77_{\pm0.014}$ \\
& & EQRM & $881_{\pm14}$ & $950_{\pm12}$ & $931_{\pm34}$ & $1627_{\pm149}$ & $1.455_{\pm0.036}$ & $1.761_{\pm0.003}$ \\
& & ERM & $875_{\pm12}$ & $932_{\pm16}$ & $996_{\pm35}$ & $1576_{\pm133}$ & $1.353_{\pm0.024}$ & $1.741_{\pm0.008}$ \\
& & GroupDRO & $875_{\pm17}$ & $934_{\pm15}$ & $993_{\pm46}$ & $1784_{\pm63}$ & $1.356_{\pm0.029}$ & $1.734_{\pm0.016}$ \\
& & IB\_ERM & $904_{\pm20}$ & $1105_{\pm60}$ & $927_{\pm15}$ & $1653_{\pm33}$ & $1.398_{\pm0.008}$ & $1.742_{\pm0.007}$ \\
& & IB\_IRM & $862_{\pm5}$ & $1136_{\pm99}$ & $911_{\pm17}$ & $1531_{\pm32}$ & $1.418_{\pm0.026}$ & $1.759_{\pm0.003}$ \\
& & IRM & $871_{\pm10}$ & $934_{\pm20}$ & $1024_{\pm14}$ & $1782_{\pm168}$ & $1.348_{\pm0.048}$ & $1.737_{\pm0.016}$ \\
& & MMD & $868_{\pm14}$ & $917_{\pm0}$ & $973_{\pm23}$ & $1579_{\pm58}$ & $1.367_{\pm0.007}$ & $1.737_{\pm0.01}$ \\
& & VREx & $942_{\pm22}$ & $1094_{\pm99}$ & $946_{\pm17}$ & $1628_{\pm49}$ & $1.349_{\pm0.02}$ & $1.739_{\pm0.003}$ \\
\bottomrule

\end{tabular}
}
\end{table}
\clearpage

\begin{table}[htbp!]
\centering
\caption{OOD generalization methods MAE on Wild-Tab benchmark with average-\textbf{out}-domain validation.}
\resizebox{0.78\textwidth}{!}{
\begin{tabular}{ll|l|ll|ll|ll}
\toprule
\multirow{2}*{Model} & \multirow{2}*{Data} & \multirow{2}*{Objective} & \multicolumn{2}{c|}{\DVpowerSynt} & \multicolumn{2}{c|}{\DVpowerReal} & \multicolumn{2}{c}{\DWeatherData} \\
& & & In & Out & In & Out & In & Out \\
\midrule
\multirow{20}*{MLP} & \multirow{10}*{Valid} & CORAL & $862_{\pm9}$ & $700_{\pm5}$ & $1008_{\pm35}$ & $618_{\pm33}$ & $1.35_{\pm0.018}$ & $1.604_{\pm0.013}$ \\
& & DANN & $988_{\pm77}$ & $704_{\pm7}$ & $973_{\pm5}$ & $632_{\pm21}$ & $1.474_{\pm0.011}$ & $1.604_{\pm0.009}$ \\
& & EQRM & $878_{\pm14}$ & $705_{\pm9}$ & $937_{\pm33}$ & $620_{\pm29}$ & $1.457_{\pm0.038}$ & $1.61_{\pm0.005}$ \\
& & ERM & $872_{\pm11}$ & $705_{\pm6}$ & $1002_{\pm36}$ & $628_{\pm10}$ & $1.357_{\pm0.026}$ & $1.603_{\pm0.008}$ \\
& & GroupDRO & $873_{\pm18}$ & $705_{\pm3}$ & $1000_{\pm47}$ & $635_{\pm44}$ & $1.36_{\pm0.03}$ & $1.599_{\pm0.004}$ \\
& & IB\_ERM & $900_{\pm22}$ & $716_{\pm7}$ & $932_{\pm15}$ & $610_{\pm5}$ & $1.397_{\pm0.008}$ & $1.62_{\pm0.007}$ \\
& & IB\_IRM & $860_{\pm3}$ & $714_{\pm4}$ & $919_{\pm16}$ & $632_{\pm53}$ & $1.418_{\pm0.025}$ & $1.623_{\pm0.008}$ \\
& & IRM & $869_{\pm8}$ & $704_{\pm2}$ & $1032_{\pm16}$ & $673_{\pm56}$ & $1.352_{\pm0.048}$ & $1.603_{\pm0.002}$ \\
& & MMD & $865_{\pm14}$ & $700_{\pm4}$ & $979_{\pm22}$ & $639_{\pm18}$ & $1.371_{\pm0.008}$ & $1.607_{\pm0.008}$ \\
& & VREx & $937_{\pm21}$ & $709_{\pm8}$ & $950_{\pm18}$ & $647_{\pm8}$ & $1.355_{\pm0.023}$ & $1.601_{\pm0.001}$ \\
\cmidrule{2-9}
 & \multirow{10}*{Test} & CORAL & $865_{\pm10}$ & $916_{\pm8}$ & $1001_{\pm33}$ & $1679_{\pm184}$ & $1.345_{\pm0.019}$ & $1.741_{\pm0.005}$ \\
& & DANN & $993_{\pm77}$ & $1029_{\pm63}$ & $965_{\pm2}$ & $1714_{\pm78}$ & $1.471_{\pm0.01}$ & $1.77_{\pm0.014}$ \\
& & EQRM & $881_{\pm14}$ & $950_{\pm12}$ & $931_{\pm34}$ & $1627_{\pm149}$ & $1.455_{\pm0.036}$ & $1.761_{\pm0.003}$ \\
& & ERM & $875_{\pm12}$ & $932_{\pm16}$ & $996_{\pm35}$ & $1576_{\pm133}$ & $1.353_{\pm0.024}$ & $1.741_{\pm0.008}$ \\
& & GroupDRO & $875_{\pm17}$ & $934_{\pm15}$ & $993_{\pm46}$ & $1784_{\pm63}$ & $1.356_{\pm0.029}$ & $1.734_{\pm0.016}$ \\
& & IB\_ERM & $904_{\pm20}$ & $1105_{\pm60}$ & $927_{\pm15}$ & $1653_{\pm33}$ & $1.398_{\pm0.008}$ & $1.742_{\pm0.007}$ \\
& & IB\_IRM & $862_{\pm5}$ & $1136_{\pm99}$ & $911_{\pm17}$ & $1531_{\pm32}$ & $1.418_{\pm0.026}$ & $1.759_{\pm0.003}$ \\
& & IRM & $871_{\pm10}$ & $934_{\pm20}$ & $1024_{\pm14}$ & $1782_{\pm168}$ & $1.348_{\pm0.048}$ & $1.737_{\pm0.016}$ \\
& & MMD & $868_{\pm14}$ & $917_{\pm0}$ & $973_{\pm23}$ & $1579_{\pm58}$ & $1.367_{\pm0.007}$ & $1.737_{\pm0.01}$ \\
& & VREx & $942_{\pm22}$ & $1094_{\pm99}$ & $946_{\pm17}$ & $1628_{\pm49}$ & $1.349_{\pm0.02}$ & $1.739_{\pm0.003}$ \\
\midrule
\midrule
\multirow{20}*{MLP-PLR} & \multirow{10}*{Valid} & CORAL & $862_{\pm37}$ & $701_{\pm7}$ & $789_{\pm32}$ & $572_{\pm24}$ \\
& & DANN & $1014_{\pm52}$ & $701_{\pm4}$ & $889_{\pm15}$ & $594_{\pm60}$ \\
& & EQRM & $1040_{\pm90}$ & $689_{\pm8}$ & $891_{\pm168}$ & $568_{\pm30}$ \\
& & ERM & $907_{\pm3}$ & $704_{\pm10}$ & $887_{\pm110}$ & $571_{\pm19}$ \\
& & GroupDRO & $780_{\pm9}$ & $705_{\pm4}$ & $769_{\pm56}$ & $578_{\pm13}$ \\
& & IB\_ERM & $880_{\pm17}$ & $697_{\pm4}$ & $918_{\pm146}$ & $618_{\pm4}$ \\
& & IB\_IRM & $915_{\pm14}$ & $700_{\pm6}$ & $926_{\pm124}$ & $618_{\pm19}$ \\
& & IRM & $878_{\pm31}$ & $699_{\pm3}$ & $853_{\pm38}$ & $620_{\pm34}$ \\
& & MMD & $968_{\pm89}$ & $702_{\pm7}$ & $843_{\pm30}$ & $587_{\pm54}$ \\
& & VREx & $982_{\pm194}$ & $699_{\pm1}$ & $778_{\pm67}$ & $564_{\pm49}$ \\
\cmidrule{2-9}
 & \multirow{10}*{Test} & CORAL & $864_{\pm38}$ & $1020_{\pm64}$ & $783_{\pm33}$ & $2327_{\pm154}$ \\
& & DANN & $1018_{\pm50}$ & $1156_{\pm68}$ & $879_{\pm16}$ & $2149_{\pm266}$ \\
& & EQRM & $1045_{\pm93}$ & $1193_{\pm69}$ & $882_{\pm166}$ & $1941_{\pm109}$ \\
& & ERM & $910_{\pm3}$ & $1080_{\pm69}$ & $879_{\pm110}$ & $2215_{\pm158}$ \\
& & GroupDRO & $780_{\pm10}$ & $1007_{\pm51}$ & $765_{\pm58}$ & $2437_{\pm302}$ \\
& & IB\_ERM & $884_{\pm17}$ & $997_{\pm12}$ & $913_{\pm142}$ & $1910_{\pm380}$ \\
& & IB\_IRM & $920_{\pm14}$ & $1018_{\pm15}$ & $919_{\pm125}$ & $2183_{\pm402}$ \\
& & IRM & $883_{\pm31}$ & $987_{\pm28}$ & $841_{\pm37}$ & $2108_{\pm195}$ \\
& & MMD & $973_{\pm88}$ & $1138_{\pm72}$ & $837_{\pm34}$ & $1945_{\pm166}$ \\
& & VREx & $988_{\pm199}$ & $1024_{\pm91}$ & $771_{\pm67}$ & $2054_{\pm436}$ \\
\midrule
\midrule
\multirow{20}*{MLP-T-LR} & \multirow{10}*{Valid} & CORAL & $844_{\pm10}$ & $691_{\pm5}$ & $698_{\pm60}$ & $520_{\pm25}$ \\
& & DANN & $841_{\pm10}$ & $695_{\pm3}$ & $1432_{\pm98}$ & $565_{\pm50}$ \\
& & EQRM & $866_{\pm41}$ & $700_{\pm9}$ & $722_{\pm47}$ & $572_{\pm87}$ \\
& & ERM & $811_{\pm8}$ & $694_{\pm2}$ & $1282_{\pm81}$ & $568_{\pm74}$ \\
& & GroupDRO & $817_{\pm5}$ & $697_{\pm4}$ & $845_{\pm127}$ & $576_{\pm16}$ \\
& & IB\_ERM & $853_{\pm9}$ & $698_{\pm1}$ & $747_{\pm53}$ & $573_{\pm19}$ \\
& & IB\_IRM & $849_{\pm33}$ & $703_{\pm15}$ & $827_{\pm23}$ & $630_{\pm15}$ \\
& & IRM & $860_{\pm38}$ & $701_{\pm1}$ & $833_{\pm60}$ & $597_{\pm16}$ \\
& & MMD & $846_{\pm11}$ & $697_{\pm8}$ & $782_{\pm56}$ & $600_{\pm51}$ \\
& & VREx & $820_{\pm11}$ & $698_{\pm1}$ & $1596_{\pm64}$ & $583_{\pm40}$ \\
\cmidrule{2-9}
 & \multirow{10}*{Test} & CORAL & $847_{\pm9}$ & $968_{\pm17}$ & $690_{\pm61}$ & $1887_{\pm250}$ \\
& & DANN & $843_{\pm10}$ & $1014_{\pm22}$ & $1420_{\pm98}$ & $2031_{\pm364}$ \\
& & EQRM & $868_{\pm41}$ & $1112_{\pm58}$ & $717_{\pm44}$ & $1898_{\pm376}$ \\
& & ERM & $811_{\pm8}$ & $1008_{\pm37}$ & $1272_{\pm84}$ & $2201_{\pm269}$ \\
& & GroupDRO & $819_{\pm5}$ & $1001_{\pm4}$ & $837_{\pm127}$ & $2611_{\pm305}$ \\
& & IB\_ERM & $855_{\pm10}$ & $991_{\pm16}$ & $741_{\pm52}$ & $2208_{\pm13}$ \\
& & IB\_IRM & $852_{\pm35}$ & $1019_{\pm39}$ & $820_{\pm23}$ & $1685_{\pm32}$ \\
& & IRM & $863_{\pm38}$ & $1003_{\pm52}$ & $833_{\pm59}$ & $1948_{\pm55}$ \\
& & MMD & $849_{\pm12}$ & $1009_{\pm76}$ & $782_{\pm57}$ & $2041_{\pm245}$ \\
& & VREx & $822_{\pm13}$ & $1021_{\pm69}$ & $1587_{\pm65}$ & $2856_{\pm109}$ \\
\midrule
\midrule
\multirow{20}*{FT-Transformer} & \multirow{10}*{Valid} & CORAL & $850_{\pm20}$ & $703_{\pm3}$ & $1097_{\pm150}$ & $586_{\pm9}$ & $1.355_{\pm0.039}$ & $1.53_{\pm0.012}$ \\
& & DANN & $891_{\pm13}$ & $708_{\pm0}$ & $1010_{\pm49}$ & $629_{\pm18}$ & $1.458_{\pm0.028}$ & $1.539_{\pm0.002}$ \\
& & EQRM & $873_{\pm59}$ & $706_{\pm3}$ & $1064_{\pm37}$ & $592_{\pm28}$ & $1.409_{\pm0.044}$ & $1.528_{\pm0.003}$ \\
& & ERM & $891_{\pm55}$ & $700_{\pm5}$ & $1048_{\pm24}$ & $582_{\pm8}$ & $1.428_{\pm0.025}$ & $1.546_{\pm0.004}$ \\
& & GroupDRO & $922_{\pm46}$ & $703_{\pm3}$ & $1025_{\pm74}$ & $604_{\pm5}$ & $1.358_{\pm0.055}$ & $1.53_{\pm0.005}$ \\
& & IB\_ERM & $933_{\pm62}$ & $704_{\pm7}$ & $984_{\pm21}$ & $589_{\pm21}$ & $1.434_{\pm0.04}$ & $1.551_{\pm0.014}$ \\
& & IB\_IRM & $859_{\pm45}$ & $711_{\pm2}$ & $1041_{\pm60}$ & $621_{\pm6}$ & $1.41_{\pm0.041}$ & $1.533_{\pm0.007}$ \\
& & IRM & $844_{\pm4}$ & $702_{\pm2}$ & $1008_{\pm38}$ & $575_{\pm9}$ & $1.482_{\pm0.006}$ & $1.544_{\pm0.004}$ \\
& & MMD & $938_{\pm56}$ & $701_{\pm6}$ & $1008_{\pm69}$ & $588_{\pm18}$ & $1.453_{\pm0.013}$ & $1.529_{\pm0.011}$ \\
& & VREx & $852_{\pm14}$ & $703_{\pm4}$ & $983_{\pm32}$ & $599_{\pm31}$ & $1.396_{\pm0.014}$ & $1.525_{\pm0.006}$ \\
\cmidrule{2-9}
 & \multirow{10}*{Test} & CORAL & $852_{\pm22}$ & $917_{\pm7}$ & $1090_{\pm147}$ & $1948_{\pm377}$ & $1.356_{\pm0.04}$ & $1.702_{\pm0.018}$ \\
& & DANN & $893_{\pm12}$ & $894_{\pm9}$ & $1005_{\pm52}$ & $2009_{\pm96}$ & $1.457_{\pm0.029}$ & $1.705_{\pm0.015}$ \\
& & EQRM & $875_{\pm60}$ & $944_{\pm49}$ & $1058_{\pm39}$ & $1878_{\pm44}$ & $1.409_{\pm0.045}$ & $1.698_{\pm0.01}$ \\
& & ERM & $889_{\pm55}$ & $967_{\pm36}$ & $1046_{\pm24}$ & $1883_{\pm130}$ & $1.427_{\pm0.025}$ & $1.724_{\pm0.003}$ \\
& & GroupDRO & $927_{\pm45}$ & $975_{\pm47}$ & $1020_{\pm73}$ & $2009_{\pm93}$ & $1.358_{\pm0.055}$ & $1.699_{\pm0.017}$ \\
& & IB\_ERM & $937_{\pm61}$ & $950_{\pm30}$ & $982_{\pm23}$ & $1877_{\pm121}$ & $1.434_{\pm0.039}$ & $1.716_{\pm0.018}$ \\
& & IB\_IRM & $860_{\pm47}$ & $903_{\pm10}$ & $1039_{\pm58}$ & $1534_{\pm190}$ & $1.408_{\pm0.043}$ & $1.716_{\pm0.004}$ \\
& & IRM & $846_{\pm5}$ & $927_{\pm18}$ & $1001_{\pm41}$ & $1789_{\pm264}$ & $1.479_{\pm0.005}$ & $1.728_{\pm0.018}$ \\
& & MMD & $942_{\pm55}$ & $1014_{\pm62}$ & $1003_{\pm69}$ & $1722_{\pm188}$ & $1.453_{\pm0.012}$ & $1.712_{\pm0.02}$ \\
& & VREx & $852_{\pm15}$ & $948_{\pm15}$ & $979_{\pm32}$ & $1653_{\pm101}$ & $1.396_{\pm0.013}$ & $1.697_{\pm0.004}$ \\
\midrule
\midrule
\multirow{2}*{catboost} & Valid & & $911_{\pm0}$ & $734_{\pm0}$ & $875_{\pm0}$ & $813_{\pm0}$ & $1.557_{\pm0.0}$ & $1.682_{\pm0.0}$  \\
& Test & & $912_{\pm0}$ & $1168_{\pm0}$ & $869_{\pm0}$ & $1719_{\pm0}$ & $1.557_{\pm0.0}$ & $1.87_{\pm0.0}$ \\
\midrule
\multirow{2}*{xgboost} & Valid & & $910_{\pm0}$ & $741_{\pm0}$ & $1002_{\pm0}$ & $934_{\pm0}$ &  $1.483_{\pm0.0}$ & $1.664_{\pm0.0}$ \\
& Test & & $915_{\pm0}$ & $1461_{\pm0}$ & $998_{\pm0}$ & $2035_{\pm0}$ & $1.482_{\pm0.0}$ & $1.828_{\pm0.0}$\\
\bottomrule

\end{tabular}
}
\end{table}

\begin{table}[htbp!]
\centering
\caption{OOD generalization methods MAE on Wild-Tab benchmark with average-\textbf{in}-domain validation.}
\resizebox{0.78\textwidth}{!}{
\begin{tabular}{ll|l|ll|ll|ll}
\toprule
\multirow{2}*{Model} & \multirow{2}*{Data} & \multirow{2}*{Objective} & \multicolumn{2}{c|}{\DVpowerSynt} & \multicolumn{2}{c|}{\DVpowerReal} & \multicolumn{2}{c}{\DWeatherData} \\
& & & In & Out & In & Out & In & Out \\
\midrule
\multirow{20}*{MLP} & \multirow{10}*{Valid} & CORAL & $794_{\pm2}$ & $803_{\pm30}$ & $623_{\pm9}$ & $954_{\pm105}$ & $1.293_{\pm0.003}$ & $1.684_{\pm0.017}$ \\
& & DANN & $819_{\pm2}$ & $751_{\pm25}$ & $766_{\pm15}$ & $925_{\pm115}$ & $1.325_{\pm0.004}$ & $1.636_{\pm0.016}$ \\
& & EQRM & $804_{\pm2}$ & $788_{\pm16}$ & $647_{\pm10}$ & $939_{\pm219}$ & $1.305_{\pm0.003}$ & $1.675_{\pm0.022}$ \\
& & ERM & $796_{\pm1}$ & $751_{\pm21}$ & $623_{\pm27}$ & $901_{\pm49}$ & $1.295_{\pm0.005}$ & $1.674_{\pm0.017}$ \\
& & GroupDRO & $797_{\pm0}$ & $760_{\pm17}$ & $611_{\pm5}$ & $1234_{\pm170}$ & $1.29_{\pm0.002}$ & $1.704_{\pm0.008}$ \\
& & IB\_ERM & $804_{\pm4}$ & $831_{\pm87}$ & $747_{\pm20}$ & $917_{\pm81}$ & $1.34_{\pm0.001}$ & $1.804_{\pm0.004}$ \\
& & IB\_IRM & $826_{\pm6}$ & $791_{\pm33}$ & $756_{\pm24}$ & $796_{\pm19}$ & $1.405_{\pm0.003}$ & $1.636_{\pm0.018}$ \\
& & IRM & $792_{\pm7}$ & $759_{\pm24}$ & $701_{\pm14}$ & $979_{\pm148}$ & $1.291_{\pm0.003}$ & $1.662_{\pm0.018}$ \\
& & MMD & $792_{\pm7}$ & $740_{\pm23}$ & $681_{\pm17}$ & $2108_{\pm841}$ & $1.295_{\pm0.002}$ & $1.671_{\pm0.009}$ \\
& & VREx & $799_{\pm0}$ & $754_{\pm27}$ & $696_{\pm9}$ & $1056_{\pm355}$ & $1.293_{\pm0.007}$ & $1.654_{\pm0.013}$ \\
\cmidrule{2-9}
 & \multirow{10}*{Test} & CORAL & $795_{\pm1}$ & $1049_{\pm43}$ & $618_{\pm8}$ & $2354_{\pm123}$ & $1.288_{\pm0.003}$ & $1.811_{\pm0.004}$ \\
& & DANN & $821_{\pm4}$ & $1042_{\pm15}$ & $759_{\pm16}$ & $1936_{\pm57}$ & $1.321_{\pm0.002}$ & $1.764_{\pm0.02}$ \\
& & EQRM & $805_{\pm1}$ & $1097_{\pm58}$ & $639_{\pm10}$ & $2188_{\pm239}$ & $1.304_{\pm0.001}$ & $1.791_{\pm0.004}$ \\
& & ERM & $797_{\pm1}$ & $1021_{\pm12}$ & $618_{\pm27}$ & $2253_{\pm375}$ & $1.291_{\pm0.003}$ & $1.803_{\pm0.007}$ \\
& & GroupDRO & $798_{\pm0}$ & $1000_{\pm46}$ & $604_{\pm6}$ & $2400_{\pm296}$ & $1.292_{\pm0.004}$ & $1.837_{\pm0.021}$ \\
& & IB\_ERM & $806_{\pm4}$ & $1084_{\pm88}$ & $740_{\pm19}$ & $2145_{\pm151}$ & $1.339_{\pm0.001}$ & $1.925_{\pm0.008}$ \\
& & IB\_IRM & $828_{\pm6}$ & $1096_{\pm37}$ & $751_{\pm23}$ & $2082_{\pm147}$ & $1.403_{\pm0.003}$ & $1.768_{\pm0.015}$ \\
& & IRM & $794_{\pm7}$ & $1039_{\pm28}$ & $696_{\pm17}$ & $2281_{\pm304}$ & $1.288_{\pm0.0}$ & $1.793_{\pm0.017}$ \\
& & MMD & $794_{\pm7}$ & $1047_{\pm90}$ & $676_{\pm16}$ & $2472_{\pm251}$ & $1.292_{\pm0.004}$ & $1.809_{\pm0.022}$ \\
& & VREx & $801_{\pm2}$ & $968_{\pm13}$ & $690_{\pm11}$ & $2299_{\pm355}$ & $1.29_{\pm0.007}$ & $1.789_{\pm0.02}$ \\
\midrule
\midrule
\multirow{20}*{MLP-PLR} & \multirow{10}*{Valid} & CORAL & $757_{\pm2}$ & $786_{\pm57}$ & $488_{\pm19}$ & $794_{\pm66}$ \\
& & DANN & $796_{\pm2}$ & $726_{\pm16}$ & $530_{\pm15}$ & $879_{\pm63}$ \\
& & EQRM & $768_{\pm0}$ & $750_{\pm31}$ & $489_{\pm12}$ & $821_{\pm110}$ \\
& & ERM & $757_{\pm0}$ & $812_{\pm14}$ & $511_{\pm26}$ & $774_{\pm79}$ \\
& & GroupDRO & $757_{\pm0}$ & $770_{\pm67}$ & $484_{\pm10}$ & $714_{\pm90}$ \\
& & IB\_ERM & $792_{\pm4}$ & $746_{\pm19}$ & $598_{\pm10}$ & $976_{\pm92}$ \\
& & IB\_IRM & $822_{\pm2}$ & $770_{\pm34}$ & $709_{\pm19}$ & $1040_{\pm130}$ \\
& & IRM & $802_{\pm12}$ & $736_{\pm10}$ & $489_{\pm15}$ & $767_{\pm103}$ \\
& & MMD & $813_{\pm14}$ & $738_{\pm26}$ & $483_{\pm9}$ & $774_{\pm28}$ \\
& & VREx & $777_{\pm4}$ & $724_{\pm15}$ & $503_{\pm11}$ & $783_{\pm20}$ \\
\cmidrule{2-9}
 & \multirow{10}*{Test} & CORAL & $756_{\pm3}$ & $1125_{\pm110}$ & $486_{\pm18}$ & $2622_{\pm244}$ \\
& & DANN & $797_{\pm1}$ & $975_{\pm28}$ & $529_{\pm13}$ & $2292_{\pm416}$ \\
& & EQRM & $767_{\pm0}$ & $1093_{\pm148}$ & $484_{\pm12}$ & $2417_{\pm469}$ \\
& & ERM & $756_{\pm0}$ & $1140_{\pm37}$ & $506_{\pm26}$ & $2538_{\pm324}$ \\
& & GroupDRO & $757_{\pm0}$ & $1055_{\pm73}$ & $480_{\pm11}$ & $2100_{\pm214}$ \\
& & IB\_ERM & $794_{\pm4}$ & $1031_{\pm30}$ & $594_{\pm11}$ & $2214_{\pm233}$ \\
& & IB\_IRM & $823_{\pm1}$ & $1027_{\pm30}$ & $707_{\pm17}$ & $6476_{\pm1534}$ \\
& & IRM & $805_{\pm13}$ & $1186_{\pm40}$ & $485_{\pm15}$ & $2564_{\pm294}$ \\
& & MMD & $817_{\pm13}$ & $1218_{\pm51}$ & $480_{\pm10}$ & $2458_{\pm370}$ \\
& & VREx & $777_{\pm3}$ & $968_{\pm77}$ & $500_{\pm10}$ & $1977_{\pm305}$ \\
\midrule
\midrule
\multirow{20}*{MLP-T-LR} & \multirow{10}*{Valid} & CORAL & $766_{\pm1}$ & $827_{\pm59}$ & $574_{\pm46}$ & $670_{\pm57}$ \\
& & DANN & $796_{\pm3}$ & $744_{\pm20}$ & $693_{\pm8}$ & $714_{\pm124}$ \\
& & EQRM & $779_{\pm1}$ & $757_{\pm9}$ & $613_{\pm10}$ & $666_{\pm109}$ \\
& & ERM & $768_{\pm1}$ & $814_{\pm66}$ & $581_{\pm16}$ & $876_{\pm59}$ \\
& & GroupDRO & $768_{\pm2}$ & $831_{\pm72}$ & $573_{\pm2}$ & $1041_{\pm471}$ \\
& & IB\_ERM & $768_{\pm5}$ & $762_{\pm33}$ & $582_{\pm2}$ & $733_{\pm134}$ \\
& & IB\_IRM & $790_{\pm7}$ & $746_{\pm46}$ & $670_{\pm3}$ & $840_{\pm137}$ \\
& & IRM & $802_{\pm3}$ & $734_{\pm6}$ & $697_{\pm8}$ & $944_{\pm65}$ \\
& & MMD & $805_{\pm1}$ & $742_{\pm25}$ & $709_{\pm22}$ & $741_{\pm53}$ \\
& & VREx & $767_{\pm2}$ & $834_{\pm123}$ & $609_{\pm14}$ & $771_{\pm117}$ \\
\cmidrule{2-9}
 & \multirow{10}*{Test} & CORAL & $765_{\pm1}$ & $1184_{\pm103}$ & $569_{\pm41}$ & $2154_{\pm89}$ \\
& & DANN & $796_{\pm3}$ & $1127_{\pm51}$ & $688_{\pm11}$ & $2152_{\pm590}$ \\
& & EQRM & $779_{\pm2}$ & $1053_{\pm21}$ & $610_{\pm10}$ & $1767_{\pm26}$ \\
& & ERM & $768_{\pm3}$ & $1124_{\pm97}$ & $575_{\pm13}$ & $1867_{\pm352}$ \\
& & GroupDRO & $767_{\pm2}$ & $1221_{\pm139}$ & $568_{\pm2}$ & $2263_{\pm154}$ \\
& & IB\_ERM & $770_{\pm5}$ & $1113_{\pm132}$ & $580_{\pm4}$ & $2046_{\pm230}$ \\
& & IB\_IRM & $790_{\pm8}$ & $1034_{\pm86}$ & $665_{\pm3}$ & $2171_{\pm312}$ \\
& & IRM & $802_{\pm3}$ & $1004_{\pm22}$ & $692_{\pm11}$ & $2092_{\pm296}$ \\
& & MMD & $807_{\pm1}$ & $992_{\pm54}$ & $707_{\pm23}$ & $2295_{\pm179}$ \\
& & VREx & $767_{\pm3}$ & $1034_{\pm55}$ & $606_{\pm15}$ & $1756_{\pm152}$ \\
\midrule
\midrule
\multirow{20}*{FT-Transformer} & \multirow{10}*{Valid} & CORAL & $787_{\pm0}$ & $754_{\pm27}$ & $719_{\pm12}$ & $929_{\pm94}$ & $1.28_{\pm0.004}$ & $1.662_{\pm0.055}$ \\
& & DANN & $820_{\pm3}$ & $733_{\pm13}$ & $910_{\pm7}$ & $714_{\pm72}$ & $1.358_{\pm0.003}$ & $1.633_{\pm0.022}$ \\
& & EQRM & $786_{\pm7}$ & $740_{\pm9}$ & $772_{\pm1}$ & $834_{\pm119}$ & $1.316_{\pm0.004}$ & $1.603_{\pm0.021}$ \\
& & ERM & $770_{\pm1}$ & $806_{\pm74}$ & $686_{\pm14}$ & $857_{\pm25}$ & $1.287_{\pm0.009}$ & $1.605_{\pm0.007}$ \\
& & GroupDRO & $774_{\pm0}$ & $833_{\pm21}$ & $711_{\pm14}$ & $840_{\pm126}$ & $1.275_{\pm0.003}$ & $1.627_{\pm0.017}$ \\
& & IB\_ERM & $820_{\pm4}$ & $761_{\pm27}$ & $830_{\pm11}$ & $1033_{\pm247}$ & $1.303_{\pm0.009}$ & $1.612_{\pm0.024}$ \\
& & IB\_IRM & $820_{\pm1}$ & $732_{\pm14}$ & $903_{\pm13}$ & $679_{\pm10}$ & $1.321_{\pm0.002}$ & $1.621_{\pm0.018}$ 
 \\
& & IRM & $815_{\pm3}$ & $730_{\pm6}$ & $911_{\pm19}$ & $696_{\pm154}$ & $1.337_{\pm0.004}$ & $1.636_{\pm0.026}$ \\
& & MMD & $817_{\pm1}$ & $745_{\pm17}$ & $869_{\pm22}$ & $898_{\pm170}$ & $1.295_{\pm0.004}$ & $1.592_{\pm0.024}$ \\
& & VREx & $767_{\pm10}$ & $788_{\pm21}$ & $708_{\pm13}$ & $857_{\pm136}$ & $1.271_{\pm0.003}$ & $1.633_{\pm0.024}$ \\
\cmidrule{2-9}
 & \multirow{10}*{Test} & CORAL & $787_{\pm3}$ & $1045_{\pm51}$ & $715_{\pm15}$ & $1873_{\pm138}$ & $1.276_{\pm0.004}$ & $1.795_{\pm0.018}$ \\
& & DANN & $821_{\pm5}$ & $926_{\pm19}$ & $906_{\pm9}$ & $2046_{\pm124}$ & $1.358_{\pm0.002}$ & $1.747_{\pm0.03}$ \\
& & EQRM & $786_{\pm8}$ & $987_{\pm45}$ & $769_{\pm4}$ & $1998_{\pm157}$ & $1.313_{\pm0.003}$ & $1.753_{\pm0.006}$ \\
& & ERM & $769_{\pm2}$ & $1077_{\pm3}$ & $681_{\pm15}$ & $2208_{\pm163}$ & $1.292_{\pm0.012}$ & $1.754_{\pm0.004}$ \\
& & GroupDRO & $774_{\pm0}$ & $1232_{\pm11}$ & $705_{\pm15}$ & $2022_{\pm108}$ & $1.274_{\pm0.004}$ & $1.784_{\pm0.024}$ \\
& & IB\_ERM & $824_{\pm4}$ & $1027_{\pm30}$ & $825_{\pm10}$ & $2001_{\pm193}$ & $1.302_{\pm0.008}$ & $1.723_{\pm0.007}$ \\
& & IB\_IRM & $821_{\pm2}$ & $923_{\pm15}$ & $900_{\pm12}$ & $1590_{\pm34}$ & $1.321_{\pm0.003}$ & $1.745_{\pm0.025}$ \\
& & IRM & $817_{\pm5}$ & $954_{\pm23}$ & $907_{\pm17}$ & $1690_{\pm163}$ & $1.336_{\pm0.001}$ & $1.744_{\pm0.017}$ \\
& & MMD & $818_{\pm2}$ & $939_{\pm26}$ & $864_{\pm20}$ & $2097_{\pm281}$ & $1.298_{\pm0.005}$ & $1.731_{\pm0.02}$ \\
& & VREx & $766_{\pm11}$ & $1130_{\pm18}$ & $704_{\pm10}$ & $2263_{\pm124}$ & $1.269_{\pm0.005}$ & $1.8_{\pm0.025}$  \\
\midrule
\midrule
\multirow{2}*{catboost} & Valid & & $809_{\pm0}$ & $758_{\pm0}$ & $728_{\pm0}$ & $913_{\pm0}$ & $1.551_{\pm0.0}$ & $1.683_{\pm0.0}$ \\
& Test & & $809_{\pm0}$ & $1188_{\pm0}$ & $721_{\pm0}$ & $1924_{\pm0}$ & $1.55_{\pm0.0}$ & $1.869_{\pm0.0}$ \\
\midrule
\multirow{2}*{xgboost} & Valid & & $767_{\pm0}$ & $790_{\pm0}$ & $572_{\pm0}$ & $969_{\pm0}$ & $1.449_{\pm0.0}$ & $1.69_{\pm0.0}$ \\
& Test & & $768_{\pm0}$ & $1622_{\pm0}$ & $569_{\pm0}$ & $2456_{\pm0}$ & $1.448_{\pm0.0}$ & $1.837_{\pm0.0}$ \\
\bottomrule

\end{tabular}
}
\end{table}

\end{document}